%% file: main.tex
\documentclass[11pt, a4paper, DIV=13]{scrartcl}
\usepackage[utf8]{inputenc}
\usepackage[english]{babel}
\usepackage{lmodern}
\usepackage[T1]{fontenc}
\usepackage[babel=true]{microtype}
\usepackage{amsmath,amsthm, amssymb}
\usepackage[
			hidelinks,
			colorlinks=true, citecolor = blue, linkcolor = blue,
            urlcolor = blue]{hyperref}
\usepackage[affil-it]{authblk}
\usepackage{url}
\usepackage{graphicx}
\usepackage{xcolor}
\usepackage{subfig}
\usepackage{framed}

\usepackage{easyReview} 
\usepackage[inline]{enumitem}	

\usepackage[natbibapa]{apacite}
\usepackage{orcidlink}

\setkomafont{disposition}{\normalcolor\bfseries}

\title{\Huge Connecting Algorithmic Fairness to Quality Dimensions in Machine Learning in Official Statistics and Survey Production}

\begin{document}

\author[1,$\dag$]{Patrick Oliver Schenk \orcidlink{0000-0003-3840-7298}}
\author[1]{Christoph Kern \orcidlink{0000-0001-7363-4299}}

\affil[$\dag$]{Corresponding author. \url{p.o.s.on.stats@gmail.com}}
\affil[1]{Department of Statistics, LMU Munich, Germany}

\date{
}

\maketitle

\begin{abstract}
\input{abstract}
\end{abstract}

\textbf{Keywords:} Algorithmic Fairness, Quality Dimensions, Machine Learning, Official Statistics, Trustworthy Machine Learning\\

\textbf{Acknowledgments:} This work has been partially supported by the Federal Statistical Office of Germany.



\newpage

\section{Introduction}	\label{sec:intro}
\input{intro}

\section{Background: Machine Learning}	\label{sec:os}
\input{ml}

\section{Background: Fairness in Machine Learning} \label{sec:fair.ml}
\input{fair.ml}

\section{Data Quality Framework Principles} \label{subsec:os.data.quality.principles}
\label{sec:os.data.quality.principles}
\input{os.data.quality.principles}

\section{Mapping Fairness to the Quality Dimensions of QF4SA}	\label{sec:mapping.fairness}
\input{mapping.fairness}

\section{Fairness Beyond the Quality Dimensions of QF4SA}\label{sec:fairness.other}
\input{fairness.other}

\section{Discussion}	\label{sec:conclusion}
\input{conclusion}

\section*{Ethics Declarations}
\textbf{Competing Interests\\}
\noindent We have no competing interests to declare.

\bibliography{references}


\section*{Appendix}	
\appendix
\input{abbrev}

\end{document}

%% file: abstract.tex
National Statistical Organizations (NSOs) increasingly draw on Machine Learning (ML) to improve the timeliness and cost-effectiveness of their products. When introducing ML solutions, NSOs must ensure that high standards with respect to robustness, reproducibility, and accuracy are upheld as codified, e.g., in the Quality Framework for Statistical Algorithms (QF4SA; \citealt{yung.et.al.2022.quality.framework.statistical.algorithms}). At the same time, a growing body of research focuses on fairness as a pre-condition of a safe deployment of ML to prevent disparate social impacts in practice. However, fairness has not yet been explicitly discussed as a quality aspect in the context of the application of ML at NSOs. We employ \cite{yung.et.al.2022.quality.framework.statistical.algorithms}'s QF4SA quality framework and present a mapping of its quality dimensions to algorithmic fairness. We thereby extend the QF4SA framework in several ways: we argue for fairness as its own quality dimension, we investigate the interaction of fairness with other dimensions, and we explicitly address data, both on its own and its interaction with applied methodology. In parallel with empirical illustrations, we show how our mapping can contribute to methodology in the domains of official statistics, algorithmic fairness, and trustworthy machine learning.

%% file: intro.tex
\paragraph{Official Statistics, Other Data Producers, and Machine Learning}
Machine Learning (ML, see Table \ref{tab:tab.abbrev} for a list of abbreviations) is now widely used in government, state, federal, and similar agencies 
(\citealp{engstrom.et.al.2020.government.by.algorithm.ai.in.federal.agencies}; \citealp{eu.nd.ai.watch, ips.nd.database.of.ai.in.public.services}; \citealp{algorithm.watch.2019.atlas.of.automation}). 
Official Statistics, e.g., in International, State, and National Statistical Organizations (NSOs for short), is one such area (see \citealp[ch.~2]{beck.dumpert.feuerhake.2018.ml.in.official.statistics} and section~\ref{sec:os}). 
The introduction of ML can be seen as part of the modernization efforts at NSOs: these happen on the (cross-)organizational level (e.g., the \href{https://statswiki.unece.org/display/hlgbas}{UNECE High-Level Group for the Modernisation of Official Statistics}
) but also within organizations because of their mandates for ongoing revision of methods, data sources, and products and, more indirectly, because of their operating principles of e.g., timeliness, and cost-effectiveness (e.g., \citealp{eurostat.2017.eu.code.of.practice}). In addition, there is increased competition from other producers of data and statistics who offer products that are, e.g., new or more timely, often made possible by gained innovation advantages or because they are less bound by quality principles \citep[ch.~2]{julien.2020.unece.hlg.mos.ml.project.report}. Thus, NSOs strive to improve by offering new or refined products (i.e., data or statistics). The latter can be described as doing better on at least one of their quality dimensions (see section~\ref{sec:os.data.quality.principles}) and not (meaningfully) worse on the others (e.g., \citealp[p.~12]{julien.2020.unece.hlg.mos.ml.project.report}): e.g., producing the `same' data more cheaply or releasing the `same' statistic more timely. 
In this endeavor, new methods, particularly ML, and new data sources, including those that require ML, are not an end in themselves but must serve the business needs of NSOs and, by extension, their audience \citep[p.~6]{measure.2020.unece.hlg.mos.ml.integration}, with demonstrated added value \citep[p.~1]{julien.2020.unece.hlg.mos.ml.project.report}. 

\phantomsection\label{intro.role.of.NSOs}
While the exact conditions for and tasks of NSOs may be laid down in local laws (e.g., the German \textit{BStatG}), by and large NSOs follow the fundamental principles adopted by the United Nations Economic and Social Council \citep{unece.2013.fundamental.principles.official.statistics} and the European Statistics Code of Practice \citep{eurostat.2017.eu.code.of.practice}: NSOs are impartial, credible producers of relevant data and of statistics, based on high professional, ethical, and scientific standards, working according to quality dimensions that we consider in section~\ref{sec:os.data.quality.principles}. The same dual roles (producers and analysts of data), a similar audience, and many of the same quality considerations are shared by (the respective data units within), e.g., central banks, federal research institutes, research data centers, units of governmental departments (e.g., the Bureau of Labor Statistics within the U.S. Department of Labor), and the big producers of administrative data such as government unemployment services, pension funds, and health insurance providers.\footnote{
    The data analyses of NSOs may tend to be more basic and descriptive.
} 
There is also great overlap with the survey world: NSOs are one of the big conductors and sponsors of surveys and have contributed much to survey methodology. Also, quality considerations for surveys and official statistics have much in common (compare \citet[ch.~2.6]{groves.surveymeth} and section~\ref{sec:os.data.quality.principles}). 
Even more importantly, several of the applications of ML in NSOs that we consider (see section~\ref{subsec:applications}) specifically concern the administration of surveys and the processing of survey data. 
Therefore, much of what we will discuss also applies to survey organizations and other data producers, so that they are included when, for brevity, we speak of NSOs.

\paragraph{Fairness}
In parallel to the introduction of ML at NSOs, the increasing use of prediction algorithms in the private and in the public sector has sparked a wide range of research on algorithmic fairness\footnote{
    No connection to the acronym FAIR (findability, accessibility, interoperability, and reusability).
}. 
We posit that algorithmic fairness is of particular relevance to NSOs as it touches on ethical considerations, quality dimensions, and their interactions. The importance of (algorithmic) bias and fairness for the work of NSOs is not completely unrecognized (e.g., \citealp{helwegen.braaksma.2020.fair.algorithms.in.context.netherlands.center.for.big.data.stats}), but treatments are sparse and more high-level. Also, fairness may get subsumed under ethics or there may be a general discussion of ethics, but not algorithmic fairness in particular (e.g., \citealt{unece.2021.ethical.considerations.of.ml}). Considering fairness solely within legal mandates and as an aspect of ethics \citep[p.~6f.]{julien.2020.unece.hlg.mos.ml.project.report} also influences the types of fairness one considers as well as, e.g., which groups are investigated. This may not be the most suitable approach for the work of NSOs. Therefore, we suggest considering fairness also within the quality frameworks of NSOs: both, as its own quality dimension and how it interacts with the other quality dimensions. In this paper, we highlight these interactions by discussing how each quality dimension of the Quality Framework for Statistical Algorithms (QF4SA; \citealp{yung.et.al.2022.quality.framework.statistical.algorithms}) maps to algorithmic fairness.

Fairness is not the only relevant dimension beyond performance \citep[p.~8]{yung.et.al.2022.quality.framework.statistical.algorithms}. Frameworks such \textit{Trustworthy ML} aim for explainable, fair, privacy-preserving, causal, and robust systems \citep{varshney.2022.trustworthy.ml.book,trust.ml.nd.homepage}.\footnote{
    There exist similar concepts and terms such as ethical or responsible computing/ML/AI. 
} 
Thus, the overlap to the quality dimensions for NSOs (section\ref{sec:os.data.quality.principles}) is very strong.

\paragraph{Contribution} 
Mapping quality dimensions of official statistics to fairness considerations leads to contributions that are of relevance to both communities. Our contribution to the literature for official statistics, survey organizations, and similar data producers includes expanding the current QF4SA framework by highlighting connections to the extensive literature on algorithmic fairness and illustrating how these connections may be exploited in practice. We thereby shed new light on the existing quality dimensions with a particular focus on how they can cater towards a safe deployment of ML at NSOs from a fairness perspective. The proposed mapping can sharpen requirements that are made in the current QF4SA -- e.g., by expanding overall accuracy assessments to notions of multi-group fairness. Nonetheless, the roles, goals, and tasks of NSOs in part differ from what the traditional fair ML literature focuses on. Thus, ML applications in NSO in turn motivate contributions to the algorithmic fairness literature: e.g., we suggest the use of heterogeneity-finding ML machinery for more fair reporting of results, for finding unfairly treated groups, and for finding biases in the data (production process). We further discuss fairness implications of (temporal) data drift, a setting that can be common with administrative data sources of NSOs, but less frequently considered in the fairness in ML community. 
Our contribution to the Trustworthy ML literature is the discussion of the interconnections among the quality dimensions, both with and without fairness.

\paragraph{Structure}
Given the heterogeneity of audiences and backgrounds, we strive for a self-contained article so that detailed background knowledge on the topics of our article is not required. Readers who are completely new to ML might want to briefly familiarize themselves with the easy-to-understand decision trees (e.g., \citealp[ch.~5.4]{molnar.2020.interpretable.ml}) as we use them as examples on several occasions. We begin by providing background on ML (section~\ref{subsec:intro.machine.learning}) and distinguishing procedural from methodological benefits of ML (section~\ref{subsec:background.machine.learning}). After illustrating the main drivers behind the interest in ML by NOS (section~\ref{subsec:drivers.goals}), we highlight ML applications (section~\ref{subsec:applications}). Next, we provide background on algorithmic fairness (section~\ref{subsec:fairness}) and the human components in fair ML (section~\ref{subsec:human}). We proceed by summarizing the quality dimensions of the QF4SA framework and their interconnections (section~\ref{sec:os.data.quality.principles}), before turning to our mapping of quality dimensions and fairness (section~\ref{sec:mapping.fairness}). Our mapping is complemented by a presentation of fairness considerations that extend beyond the dimensions of the QF4SA framework (section~\ref{sec:fairness.other}). We close with a discussion and outlook (section~\ref{sec:conclusion}). 

%% file: ml.tex
\subsection{ML and Statistics, Supervised and Unsupervised Learning} \label{subsec:intro.machine.learning}
We discuss the use of ML in NSOs in their roles as producers and analysts of data.\footnote{
    Similar to others (e.g., \citealt{beck.dumpert.feuerhake.2018.ml.in.official.statistics} and \citealt[ch.~1.2]{beck.dumpert.feuerhake.2018.proof.of.concept.ml.abschlussbericht}), we discuss neither AI that is not ML nor ML for other uses, such as virtual assistants facilitating  users' interaction with a NSO's website or data. As of our writing, chatbots powered by Large Language Models (LLMs) experience a boom and it is quite possible that they will play a role beyond improving such user interactions and experiences. The official statistics community has just started exploring LLMs -- and generative AI in general -- for its work \citep{cathal.et.al.2023.llms.in.official.stats}.
} 
Most if not all of these ML applications fall under either supervised learning or unsupervised learning.\footnote{
    We are not aware of applications using reinforcement learning \citep[ch.~10]{molnar.2022.modeling.mindsets} and we think that NSOs' work typically does not lend itself to this approach.
} 
In \textit{supervised learning}, the goal is to learn the functional relationship $f$ between inputs or \textit{features} $X_1, \ldots, X_p$ and the outcome or \textit{label} $Y$, both of which are contained in the training data. In the supervised ML paradigm, illuminated further in \ref{subsec:background.machine.learning}, the focus is on \textit{prediction}, i.e., the ability to predict the outcome from the feature values for new data points (i.e., not used for training the model): $\hat{y} = f(x_1, \ldots, x_p)$.\footnote{
    Note that the term prediction is not used in the sense of making statements about the future, although some supervised learning tasks are such forecasting tasks.
}  
In other words, the rationale for supervised ML is \textit{deployment}: being able to use the learned model $\hat{f}$ to predict the outcome for new units for which the outcome is unknown. This is in contrast with the traditional inferential statistics approach: there, the goal is the \textit{estimation} of (population) parameters $\theta$, typically in order to answer substantive questions about the world which were translated into statistical parameters, $\mathbb{E}(y|x_1, \ldots, x_p) = f(x_1, \ldots, x_p; \theta)$. Additionally, compared to prediction, population-based inferential statistics is less focused on the individual. 

Supervised learning tasks model either a qualitative outcome (called classification) or a quantitative outcome (called regression in the ML world, regardless of whether statistical regression or other methods are used). Classification predominates in the theoretical literatures (including that on fairness, see~\ref{sec:fair.ml}), in ML applications in general, and also in ML applications within NOS (e.g., \citealt[ch.~4]{beck.dumpert.feuerhake.2018.ml.in.official.statistics}).

\textit{Unsupervised learning} comprises a very heterogeneous set of methods and tasks: clustering, dimensionality reduction, and outlier/anomaly detection, but also latent variables, archetypes, association rule learning, and more \citep[ch.~9]{molnar.2022.modeling.mindsets}. In contrast to supervised learning, unsupervised tasks lack an outcome variable $Y$ in the data; only $X_1, \ldots, X_p$ are available. This also means that measuring performance is much more difficult. 
The common mission in the diverse collection of unsupervised tasks is to ``find hidden patterns'' (ibid
) or to ``discover interesting things'' \citep[ch.~12]{james.et.al.2021.intro.statistical.learning} about the units, about the features, or, more generally, about the data. It is possible that the output of unsupervised models is the endpoint of the data analysis: e.g., one may be satisfied to learn how many `groups' a clustering algorithm has detected or whether particular units are placed in the same cluster. Often, however, unsupervised methods are applied to pre-process or transform the data before they are fed into other, typically supervised models \citep[ch.~12]{james.et.al.2021.intro.statistical.learning}: e.g., a high-dimensional set of variables can be reduced to a few, more manageable, perhaps more interpretable set of `principal components' which are then employed as features in a prediction model. 
For some applications, both supervised and unsupervised learning may be useful, depending on the particular situation, goals, and available data: e.g., in the identification of the same units in two disparate data sources, one may or may not have gold-standard information about true matches via a unique identifier.

\subsection{The Machine Learning Mindset, Procedural and Methodological Benefits, and a Comparison to Statistics}\label{subsec:background.machine.learning}
\phantomsection\label{background.machine.learning}
ML-based data analysis is characterized by two somewhat separate aspects. First, the \textit{ML mindset}, paradigm, or approach which, along with its procedural contributions, we discuss in the next two paragraphs. 
Second, the increased use of \textit{ML methods} or model classes, which is discussed in the subsequent paragraph. Some model classes clearly fall under traditional statistical methods\footnote{
    By traditional or classical statistical methods, we and others (e.g., \citealt[p.~8]{dumpert.2020.unece.editing.imputation.theme.report}) do not mean historical or outdated methods but those that are part of statistical education: e.g. linear regression, but also generalized linear models, additive models, and so on. 
} 
and others are considered machine learning (e.g., decision trees, random forests, and neural networks). Thus, while there is no sharp, universally agreed-upon boundary separating the two, the notion of ML methods is still useful. 
The following discussion of why and how ML succeeds and the contributions it has brought shall be informative in several ways: to help decide whether a ML mindset and ML-based methods fit a particular application in official statistics and as necessary background information for our discussion of ML in official statistics.

With ML \textit{procedures} we refer to four practices, discussed in the next paragraph, that, empirically, are most associated with sound ML-based data analysis, but that work largely independently from whether the considered model classes are statistical or ML. At the heart of the ML mindset is the rigorous evaluation of performance on a particular task.\footnote{
    For explanations and comparisons of the different cultures and mindsets in data analysis, see \citet[ch.~2, 7, and 8]{molnar.2022.modeling.mindsets}.
} 
In supervised learning, evaluation is about how well a model is able to predict on new data, i.e., the expected out-of-sample prediction error ( \textit{generalization error}). Thus, the ML mindset is one of competition, with the best-performing model being chosen. One typically considers several model classes (e.g., linear regression, LASSO, decision trees, random forests, and XGBoost) and several models within each class. The latter correspond to different selected features and `parameters' (e.g., the features and splits in a decision tree, respectively) and different tuning or hyperparameters (e.g., the depth in a decision tree or the regularization penalty in LASSO). 

First, for both, model selection and assessment of the final model, unbiased estimation of a model's generalization error is paramount. The main procedural building block helping to ensure this is \emph{data splitting} into two parts: the training data, which are only used for training the model, and the evaluation data, which are only used to evaluate the predictive performance of the trained model.\footnote{
    In model training, this typically takes the form of repeated data splitting via cross-validation. In model assessment, there is one split into training data and test data. See \citet[ch.~7]{Hastie2009}.
} 
The error on the training data is a systematically over-optimistic measure. In contrast, the error on the separate, fresh evaluation data provides a valid estimate of the generalization error which also guards against overfitting (i.e., fitting too closely to the observed data, thus fitting partly to random noise inherent in the training observations). 
A second procedural contribution from supervised ML is that information that not, in reality, would not be available at the time of the prediction may typically not be used during model training and previous steps (see \citealp[ch.~7.8.1]{ghani.schierholz.2020.machine.learning.ml.in.big.data.and.social.science.book} and \citealp{guts.2020.workshop.on.target.leakage.in.ml.data.leakage.in.machine.learning}): \emph{Data leakage} occurs when any information from the supposedly separate, unseen evaluation data is used in some form, hurting the freshness of the evaluation data. \emph{Target leakage} is about using the values of the outcome $Y$.\footnote{
    In addition, there is a practical concern: such information may be available during training, but it would not be available when the model is actually deployed -- after all, one trains prediction model precisely because the deployment data do not contain $Y$.
} 
Both open the door for over-optimistic performance evaluations. Leakage can sneak in very subtly, as when the information is used for pre-processing the data, e.g., in feature engineering or imputation of missing data. 
Third, the centrality of performance comparisons in the ML approach brought focus to the issue of \emph{metrics} used for model evaluation and during model training (i.e., the loss function to be optimized).\footnote{
    It is possible to use a different metric during training than for model assessment, e.g., for computational reasons.
} 
This is not unrecognized in traditional statistics, 
but the choice of metrics is more active and task-driven in the ML approach.  
In classification in particular, false negatives may be much more important relative to false positives for one application than for another. Recognizing the trade-off between the two error rates and choosing a task-suitable metric is an improvement over always employing overall accuracy. 
Fourth, as the ML approach involves the consideration of several models, one natural procedural extension was to combine several models, sometimes from different model classes, into one \textit{ensemble}
(via, e.g., bagging, boosting, stacking, or simpler methods such as averaging or majority vote; see \citealp[ch.~8]{Hastie2009}). The intuition for improved predictive performance is two-fold: for different data points a different model (class) may be closest to the truth and an ensemble of models is more stable than any one model would be. 

We now turn to ML \textit{methods} or model classes and three of their reputed benefits, particularly relative to traditional statistical methods. First, flexibility, which in supervised learning is about the functional forms of the relationships between the features and the outcome as well as about interactions among the features. This actually entails two components: 
\begin{enumerate*}[label={\alph*)}]
    \item \label{itm:ability.complex.fcts} the ability to accommodate complex functional forms, which pertains mainly to quantitative features, and
    \item \label{itm:auto.recognition.fcts.and.interactions} the automatic recognition of the (approximate) functional form and of interactions.
\end{enumerate*}
The former is afforded by ML methods that can be quite complex; however, typically the more flexible, the more data are required
\citep[ch.~2.1.2]{james.et.al.2021.intro.statistical.learning}. A further reason for \ref{itm:ability.complex.fcts} lies in the ML-based \textit{approach} `trying out' many ML model classes. For a different true functional form, a different model class is the most natural fit: e.g., trees are most suitable for step functions.\footnote{
    To their credit, tree-based methods can approximate polynomial and other smooth relationships, but at the cost of increased complexity (many splits per tree or many trees in an ensemble), making them less sample-efficient for certain cases than more suitable model classes, including traditional statistical methods. 
} 
However, to compare the whole basket of ML model classes with just one statistical model class and conclude that statistics (every statistical method) is less flexible than ML (every ML method) is not fair. It also not accurate: in particular, Generalized Additive Models \citep[ch.~7]{james.et.al.2021.intro.statistical.learning} are a statistical model class that is able to automatically adapt to non-linear relationships and can accommodate interactions. In comparison studies, particularly for small and medium sample sizes, simpler and traditional statistical methods are often not inferior \citep{Christodoulou.et.al.2019.logistic.regression.not.worse.than.machine.learning,Grinsztajn.et.al.2022.tree.based.outperform.deep.learning}.\footnote{
    Also, the reported superior performance by complex or ML methods has sometimes been found to be an artifact of flawed data splitting, leakage, and other violations of the good practices discussed above (e.g., \citealp{kapoor.narayanan.2022.leakage.reproducibility.crisis.in.ML} and \citealp{roberts.et.al.2021.common.pitfalls.covid.ml.models}).
} 
Thus, even for performance reasons alone, simple methods and traditional statistical model classes should always be among those tried out; we will address other rationales such as interpretability in section~\ref{sec:os.data.quality.principles}. 
Second, automatic feature selection is built into some ML model classes: e.g., in trees, at each split, only one variable is chosen. In high-dimensional settings, feature selection helps to stabilize the model estimation and, especially for traditional model classes, is even necessary when $p \ge N$ \citep[ch.~6]{james.et.al.2021.intro.statistical.learning}. However, traditional statistics is not without methods for feature selection: e.g., subset selection procedures or, more modern, via regularization such as the LASSO \citep[ch.~3.2ff.]{Hastie2009}. 
Third, traditional statistics is geared towards what the ML culture calls structured or tabular data: e.g., for survey data represented in a matrix format, each row corresponds to exactly one respondent and each column corresponds to one survey question. \phantomsection\label{ml.deep.learning}It is undeniable that ML has made great progress regarding un- and semi-structured data such as (a collection of) images, audio or video data, texts, or even multimodal combinations thereof. In particular, Deep Learning is able to process unstructured data end-to-end: the raw input data are fed into the network -- no feature engineering needed on the part of the data analyst \citep[ch.~11]{molnar.2022.modeling.mindsets}. 

We conclude with two remarks. 
First, when the ML mindset fits an application, e.g., when prediction is the focus, then the procedural and methodological lessons discussed above are also relevant when traditional statistical model classes are used. Thus, much of the rest of this paper is not just relevant to the use of ML models. It is true, however, that more complex model classes have more potential for overfitting, so adhering to good practices tends to be more important \citep[ch.~2.1.2]{james.et.al.2021.intro.statistical.learning}. 
Second, for a quantitative outcome, the expected squared out-of-sample prediction error (ESPE) at a point $x_0$ in the feature space can be decomposed \citep[ch.~7.3]{Hastie2009}: 
$E\left( Y-\hat{f}\left( x_0 \right) |x_0 \right) = 
Var\left(Y|x_0\right) + 
\text{Bias}\left(\hat{f}(x_0)\right)^2 + 
Var\left(\hat{f}(x_0)\right)$. 
The first term, $Var\left(Y|x_0\right) = E\left( Y- f\left( x_0 \right) |x_0 \right)^2$, is the aleatoric conditional variance of the outcome around its true conditional mean $f\left( x_0 \right)$: for given features, it cannot be reduced and is independent of choices made by the data analyst.  
The second term depicts the squared bias of $\hat{f}$, i.e., the expected squared deviation of the learned model from the (unknown) true conditional mean or true model $f\left( x_0 \right)$. It is typically monotonously decreasing in model complexity, quickly at first and then leveling off \citep[ch.~2.2.2]{james.et.al.2021.intro.statistical.learning}. The greater the (allowed) complexity, the greater the set of possible models, and thus the smaller bias, i.e., the distance of the best model in the consideration set to the true model (see \citealt[ch.~7.3]{Hastie2009}). 
The third term, $Var\left(\hat{f}(x_0)\right) = E\left(\hat{f}(x_0) - E\left(\hat{f}(x_0)\right) \right)^2$, is the variance of $\hat{f}$, denoting the variation in the learned model when learned on different training data sets (same population, same sample size). In general, this estimation uncertainty is monotonously increasing in model complexity due to higher susceptibility to  small perturbations in the data \citep[ch.~2.2.2]{james.et.al.2021.intro.statistical.learning}. This is why, typically, the more flexible a model class, the more training observations are needed \citep[ch.~2.1.2]{james.et.al.2021.intro.statistical.learning}. 
Note that the implied bias-variance trade-off is due to the focus of the supervised ML approach on prediction error: any model (class), ML or not, trained to minimize ESPE will exhibit at least a small amount of bias as long as it results in a larger decrease in variance. This results in the well-known U-shaped relation of model complexity and ESPE (at least for the typical, `under-parameterized' models, see \citealt{belkin.et.al2019.reconciling.modern.ML.classical.bias.variance.tradeoff}).
ML-based data analysis operates on both sides of this trade-off: flexible model classes and trying out different models mean more complexity (low bias, high variance) whereas feature selection (more bias, less variance) reduces model complexity.

\subsection{Overarching Drivers and Goals of NSOs}\label{subsec:drivers.goals} 
The primary drivers are NSOs striving for improved products and processes (according to the quality dimensions, see section~\ref{sec:os.data.quality.principles}), new products, new applications, new data, and an interest in new methods. These are not isolated aspects: e.g., some new products (e.g., data about very dynamic sectors or companies) make only sense when released very frequently or timely and some new data are too voluminous or unstructured for existing methods (based on traditional statistics or human work). We consider two of these drivers in more detail. 

First, desired \textit{improvements} include producing data and statistics more cheaply, releasing them more frequently, more timely, or on a more granular level, making them more accurate, or lowering response burden. Partial automation is seen as a vehicle for such improvements: e.g., algorithms can handle the easy cases, allowing staff to focus on  cases that are hard to classify \citep{coronado.juarez.2020.unece.imagery.theme.report} or important or influential \citep[p.~2]{dumpert.2020.unece.editing.imputation.theme.report}, or to contribute to other activities \citep{coronado.juarez.2020.unece.imagery.theme.report}. Alternatively, algorithmic assistance can take on the form of providing a model's most likely outcomes for a given data point as suggestions in, e.g., human coding tasks 
(\citealp[p.~7]{measure.2020.unece.hlg.mos.ml.integration} and 
\citealp[ch.~7]{sthamer.2020.classification.coding.theme.report}). 

Second, \textit{new data} are considered to complement and, in part, to replace some of the traditional main data sources of NSOs -- censuses, surveys, registers, and administrative data. 
We would like to remind that survey data have already been more than just the responses to the survey items: e.g., respondents and interviewers can provide samples (soil, saliva, blood, etc.) and measurements \citep[ch.~2.2.2]{groves.surveymeth}, information from digital devices can be used \citep{keusch.et.al.wearables.sensors.apps}, and paradata about the data collection process are captured \citep{fkpara,schenk.reuss.2023.springer.volume.paradata}. How surveys will evolve in the era of, in particular, Big Data has received increasing attention since the second part of the 2010s (e.g., \citealt{baker.2017.big.data.survey.research.perspective.in.tse.book} and the \href{https://www.bigsurv.org}{BigSurv} conferences, see \citealp{hill.et.al.2019.bigsurv18}): where they can replace survey data \citep[p.~134f.]{couper.2017.new.developments.in.survey.data.collection.annual.review.in.soc}, where and how the two can complement one another (ibid; \citealt[p.~873]{japec.kreuter.biemer.lane.et.al.2015.big.data.survey.research.aapor.task.force.report}
), and what can be learned methodologically from each other \citep{hill.et.al.2021.big.data.meets.survey.science.methods}. The community appears to agree that surveys 
are here to stay: in contrast to most other data, surveys can be designed to give the desired breadth, level of detail, and fitness for a specific use, and to control the various error sources better. 
New data types may be collected in conjunction with a survey\footnote{
    This has two benefits: the survey and the other data can be designed to complement one another more optimally, and the already linked data collection makes tedious, error-prone record linkage unnecessary.
} 
or without it. Given user consent, wearables, apps, and sensors are emerging sources \citep{keusch.et.al.wearables.sensors.apps}, as are data donation and (screen) tracking \citep{ohme.et.al.2023.api.data.donation.tracking}. Instead of single values, these data exhibit complex measurement series.

Another important new data source is images -- so far mostly aerial images and other kinds of remote sensing \citep[p.~4 and 9]{coronado.juarez.2020.unece.imagery.theme.report}: in particular, there have been vast improvements in the frequency and availability, level of detail, and costs of satellite images. The volume is too much to handle for classical processes (i.e., involving traditional statistics or human work), making ML approaches a virtual necessity. Another reason is that in some cases multiple spectra or sensing technologies, going beyond wavelengths that humans can perceive, can be combined for the same object. 

Textual data are a further new avenue \citep{unece.2022.text.classification.theme.group.report}. They range from open-text responses in surveys, to traffic, coroners', or police reports, to complaint filings, building permits, and other legal documents. Some of these are acquired via web scraping, as is information from company websites, online shops, news reports, job ads, and social media posts.

\subsection{Applications and Tasks for ML in NSOs}\label{subsec:applications}
\paragraph{Before and During Collection of (Traditional) Data}
The ability to acquire representative samples depends on having high-quality sampling frames. The necessary contact and other information can come from, e.g., registers or population-wide administrative data. Automated image recognition can help in keeping the addresses up-to-date \citep{coronado.juarez.2020.unece.imagery.theme.report}, as can information scraped from company websites. The latter are also helpful for making necessary additions to and deletions from the list (e.g., new and dissolved companies, respectively). 
Duplicates on the sampling frame are another problem and they can be detected and eliminated with the help of models for \textit{identification}.\footnote{
    For two databases $\cal{A}$ and $\cal{B}$, identification has the goal to find the common units: e.g., for each record in the $\cal{A}$, it must be determined whether $\cal{B}$ has a corresponding entry and, if so, which one. In the statistical literature, this is mostly associated with record linkage \citep{herzog.scheuren.winkler.2007.data.quality.record.linkage}: e.g., for each survey respondent, one wants to identify the entry in administrative data belonging to the very same person, so as to merge the survey data and the administrative data sets. 
    Depending on the scientific field, particularly within computer science, and the specific goal, this has many different names such as entity resolution and duplicate detection. For the latter, $\cal{A} = \cal{B}$.\label{fn:record.linkage}
}\edef\fixedfootnoteidentification{\value{footnote}}\newcounter{footnoteValueSaver}\setcounter{footnoteValueSaver}{\value{footnote}}
In general, the empirics of coverage errors are understudied \citep{Eckman.2013.paradata.for.Coverage.Research.in.kreuter.paradata.book} and such new approaches are a welcome addition to the toolbox for improving sampling frames or, at least, to be able to evaluate them better.

\phantomsection\label{applications.prediction.during.data.collection}
Being uniquely suited to prediction tasks, supervised ML is the approach for forecasting (or nowcasting) what happens during data collection. Of particular interest are problems with the sampling units (likely nonrespondents, break-offs, and panel dropout) and their responses (e.g., problems understanding prompts or satisficing behavior producing subpar answers). Good predictions of these problems form the basis for interventions (e.g., via Adaptive Survey Design, see \citealp{wagner.2008.Adaptive.Survey.Design.phd.thesis}) that in turn help to increase the cost-effectiveness of the data collection and to prevent errors in the data.

Common to the mentioned tasks so far is that few features are available at the time of prediction, making paradata \citep{fkpara} and other auxiliary information attractive. \citet[ch.~5]{schenk.reuss.2023.springer.volume.paradata} provide an introduction to paradata-based applications and interventions, but mention that ML is only starting to be embraced by survey methodologists. One type of paradata are observations from the interviewers (or address listers, recruiters, or others working on the ground) about the particular dwelling and the neighborhood. Cartographic, satellite, or `Street View' information is available online but has only been modestly explored with computer vision (instead of humans) for surveys. While these data sources are in principle available upfront, they may also be outdated or unavailable for most places. We suggest that pictures are easily captured with smartphones, by interviewers or address listers, and can be processed automatically in lieu of interviewers' judgment on what to record. 

Finally, expert interviewers, especially in partly open or fully qualitative interviews, can also better prepare for visits with, e.g., web-scraped and condensed company information.

\paragraph{Processing and Adjusting Data}
\textit{Editing} is the identification of data (cells, but also variables and units) that are problematic in one of two ways \citep[p.~1f.]{dumpert.2020.unece.editing.imputation.theme.report}: Either information is missing, such as in voluntary survey responses
(e.g., working hours and experience, income, or nationality and migration background; \citealp{beck.dumpert.feuerhake.2018.ml.in.official.statistics}) or because multiple data sources were linked and a unit was not present in all of them. 
Or values are implausible, contradictory, or otherwise suspicious based on general logic, specific domain knowledge, or statistical patterns/distributions, such as survey responses suffering from satisficing or unverified parts of administrative data. 
\textit{Imputation} is the filling in of missing or the alteration/replacing of suspicious values \citep[p.~1]{dumpert.2020.unece.editing.imputation.theme.report}. 
Supervised learning on past edited data amounts to the search for the rules that govern the existing editing process: i.e., the  outcome variable for such models is whether a particular value was flagged, edited, or imputed (or non-binary variants thereof). A trained model might then come somewhat close to replicating the performance of the editing process, but should not be expected to be more accurate \citep[p.~1]{dumpert.2020.unece.editing.imputation.theme.report}. If instead true values (or some gold standard data that are better than the edited data) are available, a model trained on them may surpass the existing editing process. However, even for such data, there may be too few (documented) cases for each type of problem to be learned by supervised ML unless the mechanisms are very simple or the number of observations is enormous. Deviant interviewer behavior, up to complete fabrications, is one such example for which  unsupervised learning may therefore be the better choice  \citep{schwanhaeuser.2022.interviewer.falsification}: e.g., clustering and outlier detection.  
Finally, if the discovery of editing and imputation rules is a primary goal \citep[p.~2]{dumpert.2020.unece.editing.imputation.theme.report}, we suggest that one might also turn to the field of rule induction or (association) rule learning (see \citealp{fuernkranz.et.a.2012.rule.learning}).

In NSOs' work, \textit{outlier} or \textit{anomaly detection}, i.e., the finding of unusual or extreme data points, is typically an unsupervised task: thus, among the many methods, e.g., clustering-based algorithms exist. Data analysts usually have four choices: to ignore outliers, to remove them altogether, to impute the suspicious values, or to use robust analysis methods. In contrast, data producers can sometimes investigate the flagged data points: they may be able to confirm or correct the information.\footnote{
    E.g., if a survey respondent is listed with extreme height, the interviewer can be asked if they recall such an occurrence. In company surveys, one may contact the respondent with a request for clarification or use web-scraped or other data sources that should contain the same information.  
} 

Identification of units\hyperref[fn:record.linkage]{\footnotemark[\value{footnoteValueSaver}]} 
is a crucial step for record linkage or for the identification of duplicates. While there may be cases where supervised (machine) learning can be employed \citep[ch.~3.5.3]{tokle.bender.2020.record.linkage.in.big.data.and.social.science.book}, this is mostly an unsupervised task: in essence, for each pairing of records $a$ and $b$ from data sources $\cal{A}$ and $\cal{B}$, respectively, one wants to know their similarity in order to judge whether the belong to the same underlying unit. ML may help to implement different, data-driven distance metrics; also, as the computation of all pairwise comparisons is often infeasible, clustering or other methods may be used to replace the blocking of traditional identification units, which reduces the number of necessary operations as only units within blocks or clusters are compared.  

Nonresponse is one of the sources that can bias data. This is often countered in data analysis by employing weights that are inversely proportional to the response propensity. Predicting these response propensities is a supervised learning task (see prediction during data collection \hyperref[applications.prediction.during.data.collection]{above}). 

Some tasks can be seen as an example of both, processing and data analysis: e.g., the generation of new features. Clustering is an example of ML-based feature generation. Meanwhile, while textual data may often be fed to ML algorithms, the traditional processing steps (e.g., removing stop words, stemming, and turning text into a frequency matrix) themselves often do not involve ML.

\paragraph{Analysing Data}
ML, particularly Deep Learning, is very helpful with images. In NSOs, this has been mostly about satellite images to predict land cover and land use (agriculture, solar panels, etc.), for crop identification, monitoring of natural resources, growth of urban areas, and population distribution \citep{coronado.juarez.2020.unece.imagery.theme.report}. Such pattern recognition can be the basis for monitoring, e.g., wildlife populations \citep{bothmann.et.al.2023.automated.wildlife.image.classification} and indicators relating to climate change and the Sustainable Development Goals on agriculture, forests, and water \citep{holloway.mengersen.2018.remote.sensing.for.SDGs}. 

Many classification efforts within NSOs have been on some kind of text \citep{unece.2022.web.scraping.theme.group.report,unece.2022.text.classification.theme.group.report}: e.g., occupation coding (from open-text survey responses or job listings to ISCO or other schemes), product categories (from household spending surveys, retail sales, scanner data, or web-scraped online shop information), and classifying enterprises according to their economic (NACE) or other activities (e.g., use of AI, engagement in research and development, innovativeness, corporate social responsibility, and social media presence) 
from web-scraped website information, financial publications, and news reports. 
Causes of death, accidents, crimes, etc., can be categorized from the respective text documents.
The general classification from responses to open-ended items or transcripts in surveys is a frequent challenge.  \citep{sthamer.2020.classification.coding.theme.report}. Another task is the re-classification of past data when classification schemes are changed: e.g., historic or prior panel wave data need to be updated accordingly.

Economic and other time-series data exhibit potentially complex seasonal and other patterns which can be learned by flexible ML, given enough training data. Some of these data concern very dynamic settings (e.g., startups and high-growth firms), so prediction models can be used to extrapolate until the next data collection.

Occasionally, NSOs engage in forecasting (e.g., demographic developments) and nowcasting. GDP and other economic indicators may be released with much delay, and predicting these indicators with alternative data sources (e.g., Google Trends and traditional media information) is explored.

\paragraph{Outlook}
The applications discussed above are meant to represent the main work, goals, and interests of NSOs, but are not an exhaustive list. We conclude with some suggestions and pointers. 

Based on the user's prompts, \textit{generative AI} systems generate, e.g., new text or images. While many applications are for creative work, it could also become interesting to NSOs. First, such systems may help to generate synthetic data both in narrow the sense (released data that do not violate confidentiality) and more broadly: e.g., chatbots powered by LLMs can be used to simulate respondents or interviewers in the pre-testing phase of a new survey to detect problems. We are, however, skeptical of the idea of using LLMs as a true replacement for (survey) data collection (prompting the LLM with respondent personas and asking for the persona's response to questions one would ask in a new survey; \citealp{argyle.et.al.2023.gpt.to.simulate.human.respondents}), particularly with regard to the quality expectations of data released by NSOS. 
Second, such systems already can provide computer code for a desired task, suggesting time savings or improved quality of code and results. Such tools can make it easier for subject matter experts to interact with the data, e.g., during editing and imputation, without the need for assistance from other personnel. Time savings also imply that, e.g., more editing checks can be run. Such augmentation of human work will be particularly helpful in a data science world in which the roles of staff become blurred so that a single person has meaningful knowledge of not just one, but several or all of subject matter, statistics, ML, other methodology, and so on (\citealp[p.~6]{measure.2020.unece.hlg.mos.ml.integration} and \citealp[p.~9]{julien.2020.unece.hlg.mos.ml.project.report}). 


So far, some data collection efforts come with great response burden: e.g., documenting all household spending or all food that is consumed is very time-consuming and error-prone. With the proliferation of smartphones, respondents can simply take pictures of their receipts or prepared food. Receipts, particularly when coupled with retail data, may even offer a greater level of detail on the specific products, their prices, and so on. Such solutions are already being brought to market in the private sector.

With the power of automation, ML, and additional data, ensuring confidentiality of data and results released by NSOs is an increasingly difficult task. Yet, methods for attacking can also be used to improve defenses: e.g., generative adversarial networks to produce privacy-preserving synthetic data (see \citealp{neunhoeffer.et.al.2021.private.generative.adversarial.networks}). 

\subsection{Compatibility}\label{subsec:ml.compatibility}
We close with  a note on the compatibility of the combination of ML and traditional statistics and of data processing and eventual data analysis. We use the imputation of missing data during processing as an example. If the eventual, `downstream' data analysis is traditional, it is well known that multiple imputation, rather than single imputation, is needed to properly quantify the uncertainty of parameter estimates \citep[ch.~4f.]{little.2019.statistical.analysis.with.missing.data}. For that, multiple draws from the posterior predictive distribution are needed. A statistical imputation model provides such a distribution while ML models, even ensembles, typically do not. Conversely, suppose the eventual data analysis follows the supervised ML paradigm, i.e., prediction. If the downstream analyst is not interested in quantifying the uncertainty of the predictions, multiple imputation is not needed. However, a traditional statistical imputation model would be estimated on the whole data set while the downstream ML-user is only permitted to use the training data portion, but not the evaluation data, for learning how best to process the data (and for model training). In other words, this statistics-ML combination is likely to produce data leakage.\footnote{
    The ML-ML combination may be more likely to solve this problem by using the same data splitting.
}
The downstream ML-user may also be affected by target leakage when the outcome variable was used in the imputation model (which a downstream statistics-user typically would not view as problematic). 
These are two examples in which using ML in data processing and traditional statistics in the eventual data analysis, or vice versa, can lead to problems. 

%% file: fair.ml.tex
\subsection{Algorithmic (Un)Fairness: Sources, Concepts, and Metrics}\label{subsec:fairness}

Following controversial applications of machine learning in high-stakes settings \citep{angwin_machine_2016, Buolamwini2018, allhutter_algorithmic_2020}, fairness concerns have sparked a multi-faceted and multi-disciplinary research field centered around the social impacts of algorithmic decision-making (ADM). Research on fairness in machine learning (fair ML; see \citealt{mehrabi_survey_2021, mitchell_algorithmic_2021, makhlouf_applicability_2020, Caton2020} for overviews) is thus typically focused on prediction models as part of larger socio-technical systems which may allocate access to positions, treatments or, more generally, valuable resources. The scope of fair ML, however, extends beyond ADM applications and includes fairness implications of the use of ML in other contexts, such as in data processing and survey production \citep{rodolfa_bias_2021}.

A key concept in the fair ML literature is the notion of \textit{protected attributes}. Protected attributes are inherent or ascribed characteristics of individuals (such as ethnic origin, gender, age, or religion), for which they can (or should) not be made responsible, but which nonetheless may be the grounds for differential treatment of individuals in the real world due to prejudice and discrimination. In a narrow sense, protected attributes may be defined based on anti-discrimination legislation (such as the Equal Credit Opportunity Act in the U.S., \citealt{mehrabi_survey_2021}), but the eventual set of attributes that should be considered in a given application may be context-specific. Note that the adaptation of U.S.-centric concepts such as `race' to other contexts is also non-trivial. The implication of introducing protected attributes is now \textit{not} to ignore these features in the ML pipeline, but rather to faithfully acknowledge heterogeneity in data and to build subgroup-aware models that incorporate moral considerations on how to account for and resolve societal biases in a given context. 
To approach conceptions of fairness in machine learning, an initial, higher-level requirement could be based on the adaption of the disparate impact doctrine to data modeling -- prevent outcomes or practices that have disproportionately adverse impacts on members of protected groups \citep{barocas_big_2016}. There are various pathways through which this principle may be violated in machine learning practice, the most prominent one being (different types of) \textit{biases in data} \citep{mehrabi_survey_2021}. Historical bias may be present in any data that result from social processes: administrative labor market records capture historical discrimination on the labor market, educational attainment histories are reflective of social biases in the education system, and (geospatial) records of criminal incidents are in part affected by decisions on which areas should be patrolled. Historical bias can easily be learned by and incorporated into ML models if data that reflect social processes is used for model training. Model training may, however, also be affected by measurement bias. In supervised learning, the outcome variable that is observed in the data may be a biased proxy for the actual outcome of interest such that social biases sneak into the model in the model specification step \citep{obermeyer_dissecting_2019}: e.g., arrests are a biased proxy for criminal activity when, conditional on the same behavior, arrest probability is higher for some individuals or groups. Lastly, representation bias refers to deficits in the composition of the training data. Such deficits may refer to the (mis)representation of specific social subgroups in absolute or relative terms, or to the match between the data that is available for model training and the eventual target population more generally. We caution that very different meanings and haphazard usage of the term `representativity' in the ML/AI community have been documented \citep{clemmensen.2023.data.representativity.in.ml.and.ai}, sometimes strongly diverging from the statistical notion.
Regardless of these causes for biases in the data, there can be feedback loops: when (biased) predictions influence real-world outcomes, they may maintain or worsen biases in the next round of data, thereby sustaining or perpetuating biases in the predictions \citep{Perdomo2020}.

The fair ML literature notes that disparate impact may also be caused by other factors. This includes data pre-processing and modeling decisions along the ML pipeline which may operate next to or in interaction with existing data biases \citep{gerdon.et.al.2022.societal.impacts.of.adm.research.agenda.social.sciences, rodolfa_bias_2021}. Examples include the compilation and matching of information about subpopulations in the training data preparation step, the encoding of (correlates of) protected attributes and their use in model training, as well as decisions on how model outputs are eventually used downstream, e.g., for classification purposes. 

In light of the various ways machine learning models may be affected by social biases, an abundance of \textit{fairness notions} has been proposed in the literature which formalize different fairness conceptions and often imply corresponding \textit{fairness metrics} to quantify adherence to a given notion in practice. Fairness notions typically focus on binary classification tasks and have been formulated on the group, subgroup, or individual level. Group fairness notions compare members of protected groups to their non-protected counterparts with respect to different prediction-based quantities. Given protected attribute $A$ and predicted outcome $\hat{Y}$, independence-based group fairness notions require the predictions to be independent of group membership -- $\hat{Y} \perp A$. The separation criterion additionally considers the observed outcome $Y$ and requires independence conditionally on the true label -- $\hat{Y} \perp A \mid Y$. Sufficiency-based notions, in contrast, condition on the predictions -- $Y \perp A \mid \hat{Y}$ \citep{barocas_fairness_2019, makhlouf_applicability_2020}. Next to group fairness, subgroup fairness aims to provide stronger fairness guarantees by imposing fairness constraints on large collections of subgroups that may be defined by intersections of many (protected and non-protected) attributes \citep{hebert-johnson_multicalibration_2018, kim_multiaccuracy_2019, kearns2018}. Finally, individual fairness formulates requirements on the individual level, e.g., by mapping distances between individuals to distances in predictions (i.e., similar individuals should receive similar predictions; \citealt{dwork_fairness_2012}) or by drawing on causal reasoning \citep{kusner_counterfactual_2018, Kilbertus2017}. 

While group-based fairness metrics are rather straightforward to compute and evaluate in practice, a central result of the fair ML literature has been that `total fairness' is difficult to achieve. Except for highly stylized cases, a prediction model cannot fulfill independence, separation, and sufficiency at the same time \citep{chouldechova_fair_2016}. Requesting group fairness thus comes with trade-offs, and considerations on which (group) fairness notion should be prioritized might be highly context-specific. 

Valid criticisms of the fair ML literature must be acknowledged: e.g., some fairness notions may suggest changing a prediction model so as to provide worse predictions (for some groups) in order for some equality constraint to become satisfied. However, as discussed by \citet{kuppler2022fair}, this is an artifact of considering only ADM systems in which the decision is a function of solely the model's prediction $\hat{Y}$, ignoring the protected attribute $A$, other features used to predict the outcome $W$, and further information such as the accuracy of $\hat{Y}$ given $A$ and $W$. Such systems exist, but the flaw is in their construction, not in the consideration of the fairness of algorithms per se. This is solved by splitting up the prediction task (i.e., building a good model) and the decision-making task: then, too, fairness notions for the former and justice notions for the latter can be cleanly separated. As suggested by \citet{kuppler2022fair}, \textit{accuracy-based} or, conversely, \textit{error-based fairness} metrics may be the most natural: For classification tasks, one can ask how rates of overall errors, false positives, false negatives, 1-precision, or 1-recall, as well as miscalibration\footnote{
    Calibration requires that the predicted probabilities $p(x)$ of a ML model `mean what they say', i.e., correspond to the actual risk of observing the event that is predicted. That is, for any probability $v$, $E(y \mid x, p(x) \approx v) \approx v$.
} 
differ across groups or subgroups. For regression tasks, bias and variance can be looked at. 
This is all the more important for the work of NSOs: they do not engage in ADM, they cannot know during data production what justice principles (and other goals) downstream users of the data will have, and accuracy is already one of the main quality dimensions they consider (see~\ref{subsec:accuracy}). Furthermore, error fairness translates more easily to data that are not about humans but about, e.g., companies -- a large part of NSOs' work. 
We will therefore concentrate on error-based fairness notions later on.

\subsection{The Human Component(s)} \label{subsec:human}

The catalog of fairness notions that have been proposed in the literature highlights that fairness can be conceptualized in various (and conflicting) ways. Given a fairness metric, additional parameters might need to be set to formalize the range of values deemed acceptable. Thus, technical measures which quantify whether a prediction model satisfies some fairness constraint do not substitute for human judgment and reflection. In contrast, fair ML implies moral reasoning and raises questions of distributive justice \citep{kuppler2022fair, heidari_moral_2019, loi_fair_2021, binns_fairness_2018, lee_fairness_2020, gajane_formalizing_2018}: How should (different types of) prediction errors be distributed across social groups in a given context? Given fair predictions, which downstream allocation of resources do we perceive as just?

Committing to fairness in building and implementing machine learning systems thus requires developers and stakeholders to explicitly specify their goals. This inevitably includes engaging with various normative questions such as which attributes should be considered sensitive, which fairness concept should be prioritized, and how exactly deviations from `optimal fairness' should be defined and potentially addressed \citep{Bothmann2022}. Some guidelines have been proposed to help navigate the fairness field: \cite{makhlouf_applicability_2020} and \cite{saleiro_aequitas_2019}, for example, structure fairness notions based on a set of selection criteria. Such templates can point out critical decision points and help in guiding discussions among stakeholders, but nonetheless require normative input and context-specific weightings of interests. This implies that NSOs may need to critically engage with downstream users, and reflect on whether the same product can meet heterogeneous needs in different contexts.

Recent research has started to focus on the human component in fair ML by studying human perceptions of algorithmic fairness. This line of work focuses on how design aspects of ADM systems or characteristics of the human evaluators affect individual fairness perceptions, or how algorithmic decisions are perceived in comparison to human decision-making (see the review by \citealt{Starke2022}). Studies that investigate which type of input data \citep{Kern2022} or attributes \citep{grgic-hlaca_human_2018} are perceived as sensitive in a given context or which types of prediction errors are evaluated as particularly problematic \citep{Srivastava2019} by the general public may provide valuable input to tackle the normative dilemmas mentioned above. 

Finally, characteristics of the individual decision-maker, the algorithm, and the context in which it is applied can affect ``algorithm aversion'' or ``algorithm appreciation'', i.e., the individual's under- or over-reliance on the algorithm's results \citep{burton.et.al.2020.algorithm.aversion.literature.review,jussupow.et.al.2020.algorithm.aversion.literature.review,hou.jung.2021.reconciling.algorithm.aversion.and.appreciation}. While NSOs are typically not the place for ADM, the data they produce may very well be frequently employed for such purposes, e.g., by governmental bodies. Thus, the data and how they are produced as well as what information (documentation, metadata, etc.) is released can influence aversion to such downstream algorithms. The same can be said for the fairness perceptions discussed in the previous paragraph. 
In addition, the internal high-level decisions of whether and how to implement ML algorithms in a NSO's processes are likely affected by the very same characteristics, as are attitudes by other stakeholders (staff, recipients of statistics, data users, etc.).

%% file: os.data.quality.principles.tex
A commitment to quality is one of the fundamental principles of NSOs (see section~\ref{intro.role.of.NSOs}). Specific, lower-level criteria are required in order to concretize and operationalize this overarching goal. The European Statistics Code of Practice \citep{eurostat.2017.eu.code.of.practice} in particular contains such principles for the institutional level, for the statistical processes, and for the  outputs: relevance, accuracy, reliability, consistency and comparability (internally, over time, within and across regions), accessibility and clarity (clear, understandable, and documented), confidentiality, response burden (proportional and non-excessive), timeliness, and cost-effectiveness as `quality dimensions'. \citet[p.~1]{yung.et.al.2022.quality.framework.statistical.algorithms}, aiming to complement rather than replace existing quality frameworks, put forth a ``Quality Framework for Statistical Algorithms'' (QF4SA henceforth) consisting of five dimensions: accuracy, timeliness, cost-effectiveness, explainability, and reproducibility. The first three are visibly also part of the above list. Explainability is related to accessibility and clarity, but not fully contained within it. Reproducibility is an aspect of reliability; we agree with \citet{salwiczek.rohde.2022.quality.in.official.statistics.workshop.presentation} that robustness is another aspect of reliability and add it to the dimensions that we discuss.

\citet[p. 1 and 4]{yung.et.al.2022.quality.framework.statistical.algorithms} chose the five dimensions in QF4SA because they find them particularly relevant when ``intermediate outputs'' (that are inputs for further processing or data analysis) are produced; these dimensions, however, should also be considered upstream (relating to data and data collection) and downstream (for final statistical outputs, whether created by NSOs or external data users). We will briefly touch on to what extent these dimensions are also more relevant or different in a world with ML and therefore should be singled out. While explainability and reproducibility connect to fundamental principles, in the presented form they are sufficiently distinct from them so that they can be considered missing from previous, pre-ML quality frameworks.  

The importance of the above-mentioned quality dimensions has been established: they are central to credible, high-quality products and institutions. There is also a ML perspective on quality dimensions. \citet[ch.~2]{doshivelez.kim.2017.rigorous.interpretable.ml.iml} make the case that many problems 
stem from some form of incompleteness: models are optimized for predictive accuracy -- one important goal --, but the deployer's or decision-maker's other desiderata typically do not enter model building and training at all (as would be possible, albeit non-trivial, by introducing formal constraints or multi-objective optimization). The trained models' performance on these criteria is thus completely unknown and must be explicitly evaluated. Differences in how well models do on these quality dimensions can then be used to choose among the (similarly predictive) trained models. In any case, it must be ascertained whether the selected model fulfills minimum standards. In addition, providing explanations, limitations, and suitable applications when shipping a model is encouraged \citep{mitchell.et.al.gebru.2019.model.cards.for.model.reporting, richards.et.al.2020.creating.ai.factsheets}.

For the remainder of this section, we consider these dimensions in a ML world and mention their interconnections. The respective interactions of these dimensions with fairness are addressed in section~\ref{sec:mapping.fairness}. 
As we build on QF4SA and our remarks are complementary and typically higher-level, this chapter is best read in conjunction with \citet{yung.et.al.2022.quality.framework.statistical.algorithms}.

\subsection{Explainability and Interpretability}
\label{subsubsec:iml}\label{subsec:iml}
We begin with explainability and interpretability  which we address in somewhat more detail than the other quality dimensions.\footnote{
    We are not aware of any higher-level introductions to this topic in the respective literature on official statistics, survey methodology, and so on. Our overview is a complement to that of \citet{yung.et.al.2022.quality.framework.statistical.algorithms} who are more focused on concrete explainability methods.
}
As is widespread, we treat interpretability and explainability as synonyms (e.g., \citealp[ch.~2.1.5]{miller.2017.Explanation.in.AI.Insights.from.Social.Sciences} and \citealp[ch.~3.0]{molnar.2020.interpretable.ml}). 
Interpretability has a dual role: it is a desirable property of a model and denotes a set of tools that can help to investigate other desirable properties. 
As this dimension is not explicitly part of pre-ML data quality frameworks \cite[p.~4]{yung.et.al.2022.quality.framework.statistical.algorithms}, we give an introduction to the field of Interpretable Machine Learning (IML)\footnote{
	As ML is a subset of AI, IML should be a subset of Explainable Artificial Intelligence (XAI). Similar to how today much of AI is ML, IML is in practice not necessarily distinguished from XAI.
} 
-- on a high level, without delving into specific methods, and to the extent useful for our later discussion. We refer the interested reader to \cite{molnar.2020.interpretable.ml}'s excellent book on the subject. 

\paragraph{Concept and Background}
Interpretability can be broadly seen as the degree to which a human can understand how or why an algorithm produces its output \citep[ch.~3.0]{molnar.2020.interpretable.ml}.\footnote{
	We deliberately use the generic term \emph{output} in this definition as the common focus on \emph{predictions} is too centered on supervised ML only, although the latter is certainly the main focus.
} 
Often, this is achieved by demonstrating how inputs and outputs are related in the trained model -- globally, locally (i.e., for a specific data point), or somewhere in between (\citealp[ch.~3.0,~3.5]{molnar.2020.interpretable.ml} and \citealp[p.~5f.]{yung.et.al.2022.quality.framework.statistical.algorithms}). While many might agree with this abstract, vague conception, there is no single, universally-accepted definition of interpretability: in particular, a precise or mathematical definition of interpretability, how to measure it, and sharp boundaries are all not obvious (\citealp[p.~22071]{murdoch.et.al.2019.interpretable.ml}; \citealp[ch.~3.0,~3.4]{molnar.2020.interpretable.ml}). 

We would like to re-emphasize that what IML methods primarily do is explain a trained model. Only secondarily they also allow one to get a glimpse of (relationships and structures in) the training data and, to an even much lesser extent, of the DGP and of the `true nature of the world' -- however, all only through the narrow, often distorting lens of the trained model. 
Also, IML methods do not change a ML model: they are applied post hoc to facilitate human understanding of a trained model's behavior, but they themselves do not alter the statistical algorithm or its results in any way.\footnote{
	Of course, someone who trains an ML model might take the insights gleaned from IML methods and decide to make adaptations to the ML model. However, the IML methods themselves do not directly produce any changes.
}

Some model classes have a structure that is both, simple enough and well-understood, so that they are considered \emph{intrinsically interpretable} \citep[ch.~3.2]{molnar.2020.interpretable.ml}: e.g., the learned beta coefficients (in ML parlance: weights) of a sparse linear regression show directly how a feature's values relate to the model's predictions. Such models come with their own built-in interpretability `devices' (such as said weights), in contrast to models from the other end of the spectrum: because those exhibit high complexity and low transparency, they are considered a black box and illumination by IML methods is necessary for understanding their behavior.

\emph{Model-agnostic} IML methods work for any model class \citep[ch.~3.2]{molnar.2020.interpretable.ml}. Being able to investigate interpretability with the same IML method is key when several trained models are compared, especially when from different classes. \emph{Model-specific} IML methods can only be applied to a small set of model classes, typically because they rely on model internals that only exist for a few model classes.\footnote{
	The abovementioned inherently interpretable model classes rely on built-in IML `devices' that are model-specific: e.g., a linear regression's weights.
}

\paragraph{Scope: interpretability levels}
\begin{enumerate}
	\item \label{itm:interpret.alg.transp}
	\emph{Algorithm transparency} or \emph{mechanical understanding of the algorithm} (\citealp[ch.~3.3.1]{molnar.2020.interpretable.ml} and \citealp[p.~6]{yung.et.al.2022.quality.framework.statistical.algorithms}) is about the general, abstract knowledge of ``how an algorithm learns a model from the data and what kind of relationships it can learn'' \citep[ch.~3.3.1]{molnar.2020.interpretable.ml}.\footnote{
		E.g., linear regression fits a line through a cloud of data points so as to minimize the average squared distance from the line to the data points. Some methods such as Deep Learning are not only more complex than traditional statistical techniques but also markedly less well studied \citep[ch.~3.3.1]{molnar.2020.interpretable.ml}; to overcome their lower inherent transparency, IML is needed even more.
} 
	While such general knowledge can aid with the next two points \cite[p.~5]{yung.et.al.2022.quality.framework.statistical.algorithms}, it is completely decoupled from the specific data and the actually trained model. Consequently, it is typically not considered directly part of IML.
	
	\item \label{itm:interpret.global.alg}
	\emph{Global, model-level,} or \emph{dataset-level interpretability} 
	(\citealp[ch.~3.3.2,~3.3.3]{molnar.2020.interpretable.ml} and
	\citealp[p.~22076]{murdoch.et.al.2019.interpretable.ml}) considers how, in the trained model, inputs are related to outputs  \citep[p.~6]{yung.et.al.2022.quality.framework.statistical.algorithms}. First, on a high level, typical questions include which features were selected, which are the most important ones (by quantifying their respective contributions), and which interactions are incorporated. 
	As holding an understanding of the entire model in one's mind or visualizing it is typically beyond human capabilities, a second, modular approach is crucial \citep[ch.~3.3.2,~3.3.3]{molnar.2020.interpretable.ml}: on the feature level, the relationship of a particular feature to the output is elucidated: e.g., positive/negative/zero/non-monotone, linear/U-shaped/cutoffs/etc., moderation by interactions, and so on.
	
	\item \label{itm:interpret.indiv.predictions}
	\emph{Local, individual-level,} or \emph{prediction-level interpretability} (\citealp[ch.~3.3.4]{molnar.2020.interpretable.ml} and
	\citealp[p.~22076]{murdoch.et.al.2019.interpretable.ml}) gives \emph{explanations} of how the prediction for a particular instance (statistical unit) comes to be \citep[ch.~3.3.5]{molnar.2020.interpretable.ml}. Often, this again involves investigating how the features' values relate to the output -- but more locally than in \ref{itm:interpret.global.alg}. For instance, in binary classification: Why was the prediction `1' and not `0'? How does the output change when the value of one particular feature is altered but the instance's other feature values are kept constant? In order to receive a desired output: which feature values would need to be changed and how (typically: what is the closest (artificial) data point yielding the desired output)?\\
	Alternatively, but less frequently, the instance is contrasted with another similar, typical, or otherwise relevant data point or group of data points, whether artificial or actual.\footnote{
		The resulting comparison, however, then typically turns again its focus on the (difference in) feature values.
	}

\end{enumerate}

The separation between global, model-level (\ref{itm:interpret.global.alg}) and local, prediction-level (\ref{itm:interpret.indiv.predictions}) interpretability is useful because typically they 
use different IML methods and they have different goals and target audiences \citep[p.~22076]{murdoch.et.al.2019.interpretable.ml}. However, the boundary is not absolute \citep[ch.~3.3.5]{molnar.2020.interpretable.ml}: First, individual-level explanations can be aggregated to the level of specific groups, enabling across-group comparisons. Second, individual-level explanations can even be aggregated to the feature level. Third, the global methods can be applied to groups of instances (user-specified or formed by the model). This is important for IML methods as a tool for fairness evaluations. While fairness notions can be on the individual level, they often concern groups.

\paragraph{Outputs, Products, and Tools of IML Methods}
The types of output produced by the various IML methods are rather heterogeneous. \citet[ch.~3.2]{molnar.2020.interpretable.ml} organizes them into five partially overlapping groups. 
\begin{enumerate}
	\item Feature summary statistic: e.g., feature importance; pairwise feature interaction strengths; learned beta coefficients in linear models (which are both summary statistics and model internals).
	
	\item Feature summary visualization: e.g., partial dependence plots.
	
	\item Model internals: e.g., the
	features and thresholds used for the splits in tree-based models; learned beta coefficients in linear models.
	
	\item Data points: e.g., counterfactual data point (similar data point to a specific instance, but with the desired output; see \citealt{verma.et.al.2020.counterfactual.explanations.algorithmic.recourse.literate.review}); adversarial example (slightly different $X$ so that $\hat{Y}$ now is wrong); influential instance; prototype.
	
	\item Approximation by a surrogate model from an intrinsically interpretable model class.

\end{enumerate}

\paragraph{Considerations for Official Statistics}
First, as emphasized above, IML methods provide insights into the trained model. It is tempting to combine results from IML methods with one's own domain knowledge or intuition and believe one has uncovered some insight into the underlying DGP or the true nature of the world. However, one cannot know whether such statements are about the model or about reality. In addition, IML methods typically use only simplifications or approximations of the trained model, and different IML methods, employed to answer (seemingly or actually) the same question, sometimes provide conflicting results \citep{krishna.lakkaraju.et.al.2022.disagreement.in.XAI.practitioner.perspective}. Thus, NSOs need to be careful with respect to the nature and stability of conclusions that can be drawn from IML.

Second, many model classes used in the ML paradigm are considered black boxes. IML methods increase the transparency of such systems, increasing credibility and trust directly \citep[p.~7]{yung.et.al.2022.quality.framework.statistical.algorithms}. As IML is also employed by model developers to improve a model and by auditors to investigate it \citep[ch.~3.1]{molnar.2020.interpretable.ml}, IML usage may also increase trust in NSO's systems indirectly. Conversely, outside, pre-trained models may be harder to probe and understand, let alone fix discovered accuracy, robustness, or fairness problems.

Third, interpretability is a human and social endeavor. Characteristics of the explainer, the recipient, the (social) context, and how explanations are communicated matter and should be considered against the backdrop of human cognitive biases (\citealp{miller.2017.Explanation.in.AI.Insights.from.Social.Sciences}; \citealp[ch.~3.6]{molnar.2020.interpretable.ml}). In particular, stakeholders in different roles (e.g., model developer, model-assisted NSO staff, data user, subject of ADM, or regulator) or with different levels of subject matter or ML expertise may find different IML methods useful (\citealp{lakkaraju.et.al.2022.explainability.practitioner.dialogue}; \citealp[ch.~12]{varshney.2022.trustworthy.ml.book}; \citealp[p.~7]{yung.et.al.2022.quality.framework.statistical.algorithms}).

Fourth, interpretability is not equally important for all systems \citep[ch.~3.1]{molnar.2020.interpretable.ml}: well-understood, well-researched systems or low-stakes settings are different from high-stakes applications (e.g., ADM) or when a system is in widespread use. For foundation models, effects of algorithmic monoculture and homogenization at scale (\citealp[ch.~5.6]{bommasani.liang.et.al.2021.opportunities.risks.foundation.models}; \citealp{kleinberg.raghavan.2021.algorithmic.monoculture}; \citealp{creel.hellman.2022.algorithmic.leviathan}) have received attention; within or across NSOs, some systems will also be more important than others. 

Finally, interpretability can be of great importance in the context of specific tasks of NSOs: e.g., when the goal is to find and formalize the rules that expert annotators use to identify problematic data \citep[p.~5]{dumpert.2020.unece.editing.imputation.theme.report}. As such editing is typically accompanied by imputation, comprehensible rules might also aid in suggesting the replacement values. This application highlights the importance of choosing the set of considered ML methods: inherently interpretable decision trees, especially when combined with appropriate feature engineering, are likely to yield such editing rules, as is the field of rule induction or rule learning \citep{fuernkranz.et.a.2012.rule.learning}.

We use IML to illustrate a developing chasm between two types of (ML) data analysis: that of structured data (often with tree-based and traditional model classes) and that of unstructured data (typically with Deep Learning). For the former, we would consider, e.g.,  which features are important or, in counterfactual examples, which values someone would need to change to get a desired prediction. For the latter, features are of much less consequence, but, for images, we might highlight pixels or regions that the model relies on much or not at all and visualize them akin to heatmaps (`saliency maps'). Some `Clever Hans effects' have been discovered that way (see \hyperref[clever.hans.effects]{robustness}).

\subsection{Cost-effectiveness}\label{subsec:cost.effectiveness}
Cost-effectiveness is about the relationship between the (quality of the) outputs and the incurred costs \cite[p.~5]{yung.et.al.2022.quality.framework.statistical.algorithms}. 
Costs may be (quasi-)fixed or ongoing. Important categories include: the necessary equipment, data, and skills must be acquired; the data must be processed, a model must be trained and evaluated; equipment, skills, and models must be monitored, maintained, and updated.\footnote{
    The CO2 cost of training models, cloud storage, and so on, may not have been at the forefront so far, but will only increase in importance. Organizations, particularly those still building and changing their capacities, might be interested in `Green AI' \citep{schwartz.et.al.2019.green.ai,tornede.et.al.2022green.auto.ml,ligozat.et.al.2022.unraveling.environmental.impact.ai}. 
} 
We refer to \citet[ch.~6]{yung.et.al.2022.quality.framework.statistical.algorithms} for more details, but want to highlight some aspects. 

Standardization of processes is one important tool to manage cost-effectiveness and other quality dimensions
(e.g., \citealp[Indicator~10.4]{eurostat.2017.eu.code.of.practice} and \citealp{destatis.2021.quality.manual}). Automation, driven by ML (or statistical) models, is a promising avenue in this regard. One anticipated benefit is that automated processes may entail higher (quasi-)fixed costs of setting up -- e.g., for equipment, knowledge acquisition, and the training, selecting, and evaluating of models -- but once they are implemented, the marginal cost per additional unit -- e.g., the cost to generate a prediction for an additional data point -- is very low: i.e., such processes are highly scalable.

We consider two cost aspects in more detail. 
First, the work does not stop with training a model: it must be evaluated on more than its performance -- namely its interpretability and its fairness -- and it must be continuously monitored after deployment for model drift or decay (declining performance and other changes in behavior) and, when needed, re-trained; this binds manpower, computational resources, and may necessitate further data and data processing work
\citep[ch.~1 and 4]{ml.group.2022.model.retraining}. Note that these requirements are largely independent of whether the chosen model for automation is ML or statistical: changes in real-world mechanisms or in data collection affect them both. Choosing models that are more stable (see~\ref{robust.fairness.causal}) may thus provide financial relief via a decreased need for re-training.  
Second, data are not only at the core of model performance but also an important consider cost consideration. ML models and the ML paradigm typically exhibit high demands regarding computational resources and data volume,\footnote{
    There is, however, a notion of `tinyML' -- which can have the additional benefits of being able to run on, e.g., respondents' smartphones so that confidential information may be processed on the device and never leave it.
}
affecting costs, timeliness, and the uncertainty (\ref{subsec:accuracy}) of the resulting output. The flexibility of ML is one of its advantages, but also increases data requirements: the less `known' structure and other types of relevant expertise are used, e.g., to create and transform features, the more the method must learn on its own -- Deep Learning on unstructured data is the prime example. 
Also, the training data for supervised editing and imputation models need to be carefully labeled, requiring perhaps more time than the simple editing and imputation itself would, and, if anything changes, existing training data may need to be re-relabeled  \citep[p.~6]{sthamer.2020.editing.social.survey.data.with.ML}. 

Published cost studies are rare in general and findings for one setting or organization may not translate directly to another (see for \citealp[ch.~5.3.6]{groves.surveymeth} on survey costs). ML and automated solutions have not shown to be always more cost-effective for NSOs' applications, but there are positive examples \citep[p.~13]{sthamer.2020.classification.coding.theme.report}. In addition, switching from one process to another is not cost-free. So far, in applications such as editing and imputation, no single ML method clearly dominates and a lot of work may be required, especially to yield more than marginal benefits, and a full range of model classes must be prepared and considered for each application \citep[p.~7]{dumpert.2020.unece.editing.imputation.theme.report}. 

Unsurprisingly, automation is also being pursued in the ML world. First, automated machine learning (AutoML; see \citealp{tornede.et.al.2022green.auto.ml,weerts2023}) is concerned with automating the whole pipeline, from data pre-processing and feature engineering to model training, hyperparameter optimization, and model selection. 
Second, monitoring of a deployed model for drift \citep{ml.group.2022.model.retraining} can also be automated. Yet, complete automation is not the goal of NSOs, and expertise and skills are still needed to implement and monitor these even more automated systems. This is also evident in Deep Learning: being able to process (raw) data end-to-end, it may not require (costly) feature engineering, but choosing the proper types of neural networks, optimizing its building blocks, and choosing the best (hyper)parameters does require expertise and some work (e.g., \citealp[p.~7]{coronado.juarez.2020.unece.imagery.theme.report} and \citealp[ch.~10]{james.et.al.2021.intro.statistical.learning}).

\subsection{Timeliness}\label{subsec:timeliness}
Timeliness can be broadly seen as ``the time between a need [...] and the release of the information to meet that need''; particularly for information covering a certain point or period of time, this is often conceptualized as the time between that reference point or period and when the information is made available \citep[ch.~5]{yung.et.al.2022.quality.framework.statistical.algorithms}.\footnote{
    Timeliness is also about the punctuality of outputs. We will not refer to this explicitly other than by mentioning that when the need to re-train a drifted model arises, this might cause delays, particularly when the issue is discovered late and there is no buffer.
} 
Economic indicators such as GDP and inflation are examples of time-sensitive information: the more delayed they are released, the less relevant and valuable they are to decision-makers. For the work of NSOs, once should consider time for data collection and acquisition, for data processing, and for data analysis. NSOs have processes for these three tasks and \citet[ch.~5.2]{yung.et.al.2022.quality.framework.statistical.algorithms} differentiate between the time needed for the development of a process, i.e., from conceptualization to implementation, and the time needed for the process to run. These processes can be sequential so that one cannot begin before the previous one is finished: bottlenecks, such as editing and imputation, may thus particularly benefit from improved, model-based processes \citep[p.~5]{dumpert.2020.unece.editing.imputation.theme.report}.

A different aspect of timeliness is the ability of a model to be used in (near) real-time, particularly to assist with data collection. In surveys, models may be employed to predict the likelihood of break-offs or of poor answering behavior and intervene accordingly (e.g., \citealp[ch.~6]{mittereder.2019.predicting.preventing.breakoff.web.surveys.phd.thesis}). They may also be used to evaluate data accuracy\footnote{
    Such evaluations may use ML and they may use multiple data sources, including new data sources mentioned above to check, e.g., survey responses or information provided by interviewers.
} 
as interviewers or respondents on the spot should be more able to correct errors than data processing staff can do later on. Sophisticated models that would need constant re-training during ongoing data collection might be too slow to be implemented.

\subsection{Robustness}
\label{subsec:robustness.stability}
\label{subsubsec:robustness.stability}
\citet{salwiczek.rohde.2022.quality.in.official.statistics.workshop.presentation}, adding this dimension to the QF4SA framework, define it broadly: ``slight perturbations of circumstances should only lead to (at most) slight changes in the output'' (words in parentheses added by us). There is some similarity to reproducibility (see~\ref{subsec:reproduc}), but robustness is more concerned with the sensitivity to \textit{small} changes 
than reproducibility (completely new method or data). Importantly, different communities have different understandings of `robustness' that overlap only a little. 

First, in the ML community, \textit{adversarial robustness} is about the ability to withstand attacks \citep[ch.~11]{varshney.2022.trustworthy.ml.book}: adversaries may either target the modeling phase, poisoning the training data by injecting additional data or by modifying data in order to change the trained model's behavior (e.g., generally lower accuracy or a different prediction for specific, targeted points in the feature space); or they may target the deployment phase to be able to evade the model's `intentions', ``gaming the system'' \citep[ch.~3.1]{molnar.2020.interpretable.ml}. While `attacks' might sound overt, documented examples include how slight, imperceptible-to-humans changes to some of an image's pixels can change classification drastically. When NSOs collect their own data, there are typically very few, if any units that have the capacity to modify the training data meaningfully (very large companies might be a counterexample). This is different when there is reliance on outside data and data providers. While not used by NSOs themselves for decision-making, governmental institutions and others may base their decisions on data or results provided by NSOs, making the topic of evasion not completely irrelevant.

Second, robustness in the ML community can also be in regard to \textit{data shifts} and \textit{model drift}, i.e., to changes between training and deployment (\citealp{quinonero.candela.et.al.2008.dataset.shift.in.ml.edited.volume}; \citealp[ch.~7]{morreno.torres.et.al.2012.dataset.shift}). This an aspect of \textit{transportability}, i.e., the question of whether a model trained on one data set also holds, without bias, for a different set of circumstances: In the supervised ML paradigm, the main concern is whether the target population (deployment data) and the training population (training data) exhibit the same distribution or whether there is a shift. Traditional inferential statistics is mostly worried about whether estimates from the data generalize to a `general population'.\footnote{
    An exception: the question of whether measurement error models estimated on one gold-standard data set can also be transported to another data set \citep[ch.~2.2.4]{carroll.et.al.2006.measurement.error}.
}\phantomsection\label{explain.transportability} 
Transportability is a more general concept than these two concerns: any change in the circumstances, environment, or context may pose a threat to the validity of a model outside its training data. Among these, the TSE framework highlights changes to the data collection protocols and data processing procedures.

Third, traditional statisticians might be inclined to think of \textit{robust statistics}: the ``insensitivity to small deviations from the assumptions'' of models in actual, finite data 
\citep[ch.~1.1]{huber.ronchetti.2009.robust.statistics} -- in particular, the robustness of the results to the presence of outliers and otherwise extreme data points (possibly the result of gross errors) in the data (see also \citealp[ch.~1.1]{hampel.et.al.1986.robust.statistics}). Ensembles are a ML answer to a lack of robustness of individual learners: this is the advantage of, e.g., random forests over a single tree \citep[p.~340]{james.et.al.2021.intro.statistical.learning}. Ensembles can be specified according to fixed rules such as averaging or majority vote. It is also possible to learn how best to weight an ensemble's components. Clustering is another application that is often prone to instability. It can also demonstrate another way to employ ensemble thinking: different cluster models can be compared regarding their conformity and a robust combined model can be created that only contains results that are common to many or even all of the models \citep{hornik.2005.clue.explaination}. 

Robustness does not only concern the model itself but also the assessment of its accuracy (\ref{subsec:accuracy} and model interpretations (\citealp[ch.~3.5]{molnar.2020.interpretable.ml}; \citep{dutta.et.al.2022.robust.counterfactual.explanations}). Conversely, IML methods can uncover which features are important in a model. These may then be more closely monitored for signs of drift. Particularly for the adversarial and drift notions of robustness, features that exhibit a causal effect on the outcome are often preferred over mere correlative features (e.g., \citealp[ch.~3.1]{molnar.2020.interpretable.ml}): causal relationships may be much more stable over time and much harder to manipulate or game. A model's most important feature can be found with IML and investigated in this regard. Similarly, if the most important features are just proxies, we might improve future data collections to come closer to the actual variables of interest, increasing statistical robustness by reducing measurement error: e.g., survey questions may be tweaked or alternative data sources considered. \phantomsection\label{clever.hans.effects} Finally, markedly worse than spurious features are so-called shortcut learning and Clever Hans effects: i.e., when the training data contain signals about $Y$ that in deployment will not be present or exhibit a very different relationship \citep{bellamy.hernan.beam.2022.shortcut.features}. Examples in medical image analysis for disease prediction include the type of imaging device, image timing, watermarks, or, worst, circles or arrows pointing to tumors that were of course not present on the disease-free among the training images (e.g., \citealp[p.~104f.]{chen.et.al.2019.ai.in.medicine.potential.and.shortcut.learning}).

\subsection{Reproducibility}	\label{subsubsec:reproduc}\label{subsec:reproduc}
Within and across scientific disciplines, a number of contrasting definitions and terminologies exist in this area \citep{barba.2018.terminologies.for.reproducible.research,
plesser.2018.reproducibility.replicability.history.of.terminology}. They all describe under which conditions the results from a new data analysis must be the same or qualitatively similar to those from a first data analysis \citep{goodman.et.al.2016.reproducibility.definitions}: 
\textit{Methods reproducibility} is defined as obtaining identical results, using the \textit{same data} and the \textit{same methods}. By leaving the era of `point-and-click adventures' and instead archiving code and data and making them accessible, scientific communities approach what is increasingly seen as the bare minimum for credible empirical work. The importance of metadata and proper documentation has also been emphasized \citep[ch.~5]{ml.group.2022.model.retraining}. 
\textit{Results reproducibility} is about finding the same results using \textit{different data} but the \textit{same methods}. 
We consider \textit{inferential reproducibility} as whether (qualitatively) the same results are produced with the \textit{same data} but \textit{different methods}.\footnote{
    In these definitions, `methods' are to be understood broadly, with analytical choices big (e.g., the considered model classes) and small (e.g., options in an algorithm implementing a model class), but also how to proceed with, e.g., outliers. Also, implied in these definitions is that all data analyses concern the same research question or research object \citep{goodman.et.al.2016.reproducibility.definitions,salwiczek.rohde.2022.quality.in.official.statistics.workshop.presentation}.
} 

Reproducibility is a big part of Open Science and ``Open Science is just good science in a digital age'' \citep{seibold.2023.quote.open.science.good.science.digital.age}, regardless of the type of data analysis. Thus, this important quality dimension is not tied directly to ML. It is, however, true that more sophisticated model classes are more likely to contain stochastic elements; also, data splitting in the supervised ML approach introduces randomness. If uncontrolled, these random elements are a threat to \textit{methods reproducibility}. Yet, this is not restricted to ML: e.g., traditional Gaussian clustering methods may use random initialization values. 
Versioning, referencing, and archiving of code and data are relatively straightforward for a traditional data analysis world where everything is in-house. However, new data sources may be non-static (e.g., even \textit{past} social media platform data are frequently changed retroactively, see \citealp{west.wagner.2023.tdq}) or too big. A similar argument pertains to large, pre-trained models (e.g., foundation models) trained and provided by an outside organization and employed by a NSO possibly after some fine-tuning. Any process containing human decision-making is more difficult to archive and to reproduce than an automated one, although very strict, documented guidelines may help. 

While we agree with \cite[p.~20]{yung.et.al.2022.quality.framework.statistical.algorithms} that NSOs often cannot easily collect new data -- especially if they are to be collected in precisely the same manner --  we, however, do not believe that this makes \textit{results reproducibility} generally unachievable or irrelevant to NSOs. In fact, whether new data sources permit the same results (but more cheaply or timely, see~\ref{subsec:drivers.goals}) is directly coupled with 
questions of results reproducibility. 
Also, recall that the ML paradigm is typically not about just analyzing one data set to answer questions: rather, the rationale is the deployment of the trained model on new data -- often more than once or even continuously. How well the model holds up is about results reproducibility and the reason for monitoring for model drift.

Finally, we note that `results' in the above-mentioned reproducibility definitions are understood to refer to outputs of data analyses. NSOs as producers of data that are used, often in multiple ways, by end users inside and outside the organization, may also consider the reproducibility of data processing (or data production more generally). This is also true for additional information they release: e.g., IML methods may contain stochastic elements.

\subsection{Accuracy}
\label{subsec:accuracy}
The many conceptions and measures of accuracy share a common notion \cite[ch.~3]{yung.et.al.2022.quality.framework.statistical.algorithms}: accuracy is about the closeness of `what one has' to the truth or, when `truth' is not an adequate concept, to what one intended.\footnote{
    For instance, when subjective opinions of a survey respondent are sought, `truth' might not be the best concept.
} 
Inspired by NSO's dual role as producers of data and of statistical outputs, we think of three kinds of objects of interest `which one has': the data, estimates of some population parameters (in traditional statistics), or predictions (in the supervised ML paradigm). In particular for the data and the predictions, one can take an individual view (a specific data point or a local prediction $\hat{y}|x_0$) or an aggregate view (differences in the distribution of the data relative to that of the target population or quantifying a model's overall performance with one accuracy number).

Statisticians tend to think of the deviations from the truth/intention as either systematic (bias) or random (variance). Survey methodologists often employ the Total Survey Error (TSE) framework to conceptualize the different sources for such errors \citep[ch.~2]{groves.surveymeth}: errors of measurement (i.e., deviations of the values in the cells of the data matrix from the truth), errors of representation (i.e., differences in the composition of the analyzed data relative to the target population), and errors occurring during data analysis -- although the latter are often not explicitly considered 
in TSE-based operations. Whether data analysis employs ML or not, the TSE framework remains a powerful tool for planning data collection and considering data quality \citep{unece.2022.quality.of.training.data.theme.group.report}. Yet, we suggest that the extensions of the TSE to newer data sources (e.g., Big Data, see \citealp{amaya.biemer.kinyon.2020.total.error.framework.big.data.from.tse}) or to data sources more general than surveys \citep{west.wagner.2023.tdq} may provide additional value. 

Accuracy metrics are at the core of the training, evaluation, and selection of models in the supervised ML paradigm (see~\ref{subsec:background.machine.learning}). For the selection in particular, two types of comparisons exist. First, the relative comparison of models, i.e.,  against each other, 
is used for model selection. If NSOs wish to test new procedures involving ML against existing procedures, both must be evaluated on equal ground: ideally, on the same unseen evaluation data and in a manner identical to what the actual implementation in practice would look like. This is also true for ML-assisted procedures, e.g., combining a model's results and human work. 
Second, there is also an absolute comparison of a (chosen) model's accuracy: how well does it perform? Does it achieve a required minimum standard? We want to highlight a common problem in the discussion of (binary) classifiers: often, a high overall accuracy (i.e., the proportion of correct predictions $\hat{y}=y$) such as 0.91 is touted as evidence for a great model, implying that the suitable reference point might be 0.50 or 0. 
However, the correct reference point is the frequency of the majority class -- and this piece of information is often missing from performance discussions. To see why it is crucial, consider an imbalanced classification problem in which the majority class occurs with a frequency of 0.9. The simple model containing only a constant and hence always predicting the majority class thus has an accuracy of 0.9.\footnote{
    Suppose that the majority class is the `positive' class ($y=1$). The constant model then has a true positive rate or sensitivity of 1.0, as it predicts only positive labels, and a true negative rate of specificity of 0. It can be shown that its accuracy is equal to the frequency of the majority class.
}
Suddenly, the added predictive ability of $0.01 = 0.91 - 0.9$ achieved by the selected model and its features is recognized to be only tiny.\footnote{
    Note the contrast to regression. A regression model that only contains the intercept has, by definition, $R^2 = 0$, and for a model containing only irrelevant features one would expect $R^2 \approx 0$ so that 0 is indeed a valid reference point.
} 
Alternatively, as in section~\ref{sec:mapping.fairness}, one may use \textit{balanced accuracy}, i.e., the unweighted average of the true positive rate and the true negative rate, for which the constant classifier always achieves a value of 0.5.

Accurate results hinge on accurate, i.e., high-quality training data (e.g., \citealp[p.~8]{coronado.juarez.2020.unece.imagery.theme.report}). Deviations might come in (or be remedied) at any level described by the TSE framework. Their consequences depend on the type of data analysis. If prediction is the ultimate goal, then the error mechanisms in the training data should be as close as possible to those in the deployment data. This is another example of transportability (see section \ref{explain.transportability}).  
If, however, the data are analyzed with traditional statistics, any information about the error components and mechanisms is helpful for deriving unbiased estimates via, mostly, measurement error models or mixed (hierarchical, multi-level) models. In survey data, the contributions at different levels are acknowledged, e.g., via fixed or random effects on the level of respondents, interviewers, and items (e.g., \citealp{couper.kreuter.2013.response.times.jrssa}). Yet, similar information about data processing is often not released: e.g., who annotated a particular data point -- a ML model or a human (and if so, a pseudo-id for the particular annotator). We must acknowledge that research on how to optimize guidelines, instructions, and other characteristics for annotation tasks is still nascent (\citealp{beck.et.al.2022.improving.labeling.through.social.science.research.agenda}; but see, e.g., \citealp{fort.2016.collaborative.annotation.book} on annotating texts).

Accuracy not only concerns the outputs (estimates or predictions) but, especially in the ML paradigm, also the performance evaluations. Adherence to good practices documented in section \ref{subsec:background.machine.learning} is key, but violations are not necessarily obvious. In particular, any type of initial, exploratory data analysis (influencing feature engineering) and kind of data processing should only be done on the training data, not the whole data including the evaluation data, in order to prevent data leakage and overoptimistic performance evaluations. Target leakage should also be avoided -- but this is difficult when data processors do not know the eventual data analysis, i.e., they do not know which variables are outcomes. 

Particularly for results of data analysis released by NSOs, uncertainty assessments are required \citep[p.~13]{yung.et.al.2022.quality.framework.statistical.algorithms}. While traditional statistical methods come with  `self-assessed' uncertainty quantifications, some ML model classes do not (e.g., support vector machines) while others do (e.g., random forests). For classification tasks, the predicted class probabilities are an uncertainty measure; however, ML model classes typically do not quantify how uncertain these probabilities themselves are.\footnote{
    Class probabilities express aleatoric uncertainty: that which is caused by the randomness inherent to the non-deterministic relation $y|x = f(x)$ depicted by the `true model' $f$. The uncertainty about the predicted probabilities is epistemic: one does not know how the trained model $\hat{f}$ deviates from the truth $f$. 
    See, e.g., \citet{bengs.huellermeier.waegeman.2022.difficulty.of.epistemic.uncertainty.quantification}.
} 
One way to express uncertainty is, instead of point predictions, to output
prediction sets (for multi-class outcomes, e.g., many image classification tasks) or prediction intervals (for continuous outcomes): 
conformal prediction (e.g.,  \citealp{angelopoulos.bates.2022.gentle.intro.to.conformal.prediction}) is a technique to turn predicted probabilities or scores into such sets or intervals -- with desirable properties even when the underlying model is not perfect.\footnote{
    Note that prediction intervals of traditional statistics are based on the assumption of having specified the model correctly.
} 

Note that the ML performance comparisons typically do not involve the uncertainty inherent to having to estimate the models' accuracy. This is unsatisfactory for NSOs: e.g., changing an existing process to a new, ML-based procedure is not cost-free and an organization wants to have some level of confidence about what procedure it should choose. 

%% file: mapping.fairness.tex
In the QF4SA, fairness considerations are, at best, discussed as a secondary aspect, e.g., in the context of explainability. As the frameworks' main focus on ``intermediate outputs'' contrasts with the typical ML use cases that are discussed in the fair ML literature, this missing link may not be surprising. However, as we argue in the following sections, bringing in a fairness perspective, both conceptually and in practice, is critical for a wide range of ML applications, particularly including the uses of ML that are (prospectively) prominent at NSOs (see section \ref{subsec:applications}). The interactions between algorithmic fairness and the QF4SA further contribute to the existing quality dimensions by highlighting blind spots and introducing methodology that targets explainability, reproducibility, robustness, and accuracy from a different angle.

\paragraph{Empirical example} We make use of a machine learning application for algorithmic profiling in the public sector \citep{kortner2021, desiere_statistical_2019} to illustrate how fairness considerations may be mapped to the QF4SA. All models that are presented are based on data from German administrative labor market records, concretely the \textit{Sample of Integrated Employment Biographies} \citep[SIAB,][]{siab2019} maintained by the Research Data Center of the German Federal Employment Agency at the Institute for Employment Research (IAB). The data include information on (un)employment histories of job seekers for the period between January 1, 2010 and December 31, 2016. The prediction task is to classify, at entry into unemployment, whether an unemployment episode will last longer than one year (long-term unemployment; LTU). For more details see \cite{kern2021}.

\subsection{Explainability \& Interpretability}
Explainability and fairness can be viewed as strongly intertwined processes throughout the ML pipeline. At the development stage, IML methods can help understand whether and how a model inherits societal biases. To this end, initial steps may include investigating the role and importance of (correlates of) protected or sensitive attributes and studying whether `legitimate' features are utilized in different ways for social subgroups. At the deployment stage, the (perceived) degree of interpretability may shape fairness perceptions of the eventual `user' of the algorithm, and their reliance on the model's outputs. If IML methods are used in either the production or deployment stage, another consideration is the degree to which the IML methods' \textit{fidelity} varies by group:\footnote{
    Recall that IML methods often involve approximations of the actual prediction model. Fidelity is the correctness or accuracy of how an IML method describes the model's behavior \citep[ch.~3.5]{molnar.2020.interpretable.ml} -- crucial to the method's value.
} 
if the explanations are not able to correctly reflect the models' decisions similarly across the feature space, any conclusions that are drawn about the models' functioning can be differently accurate across subgroups. 

\begin{figure}[h!]
\centering
\subfloat[Surrogate tree for Germans]{\includegraphics[scale = 0.425]{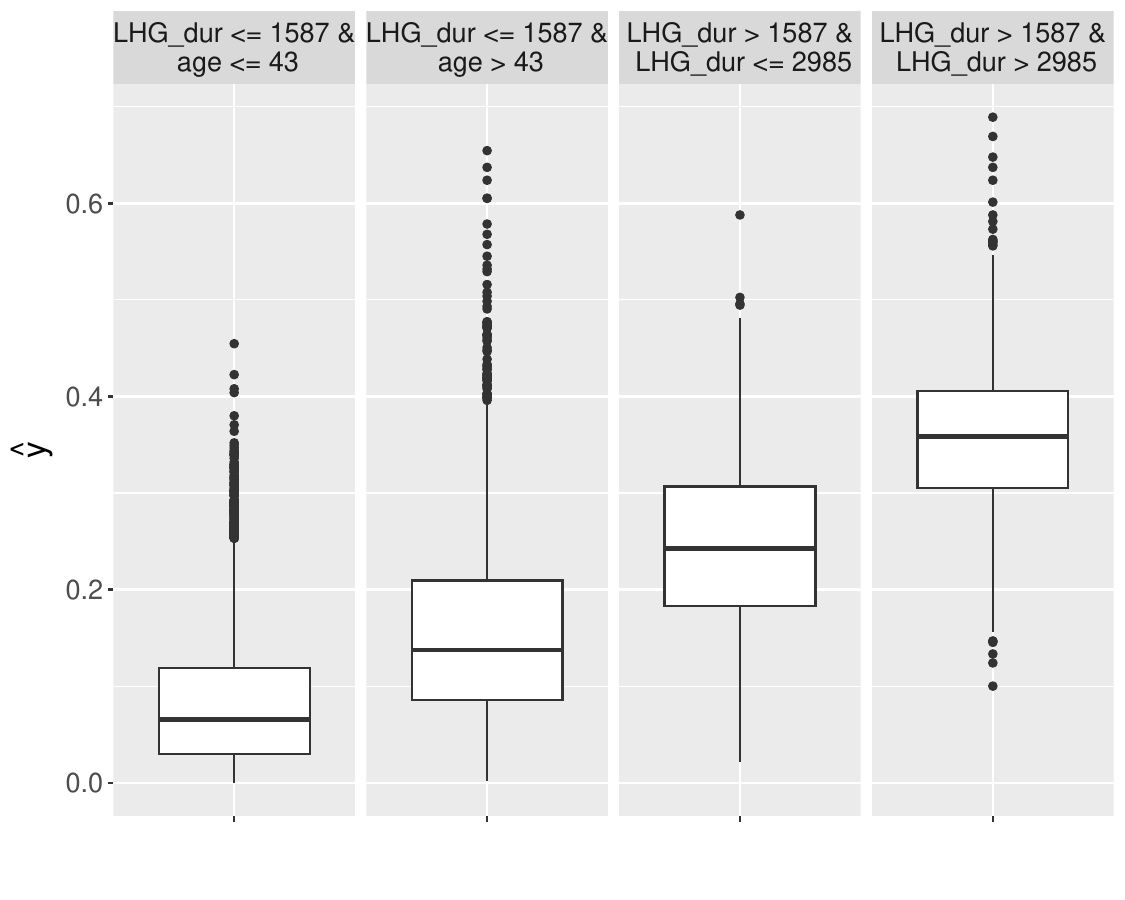}}%
\subfloat[Surrogate tree for non-Germans]{\includegraphics[scale = 0.425]{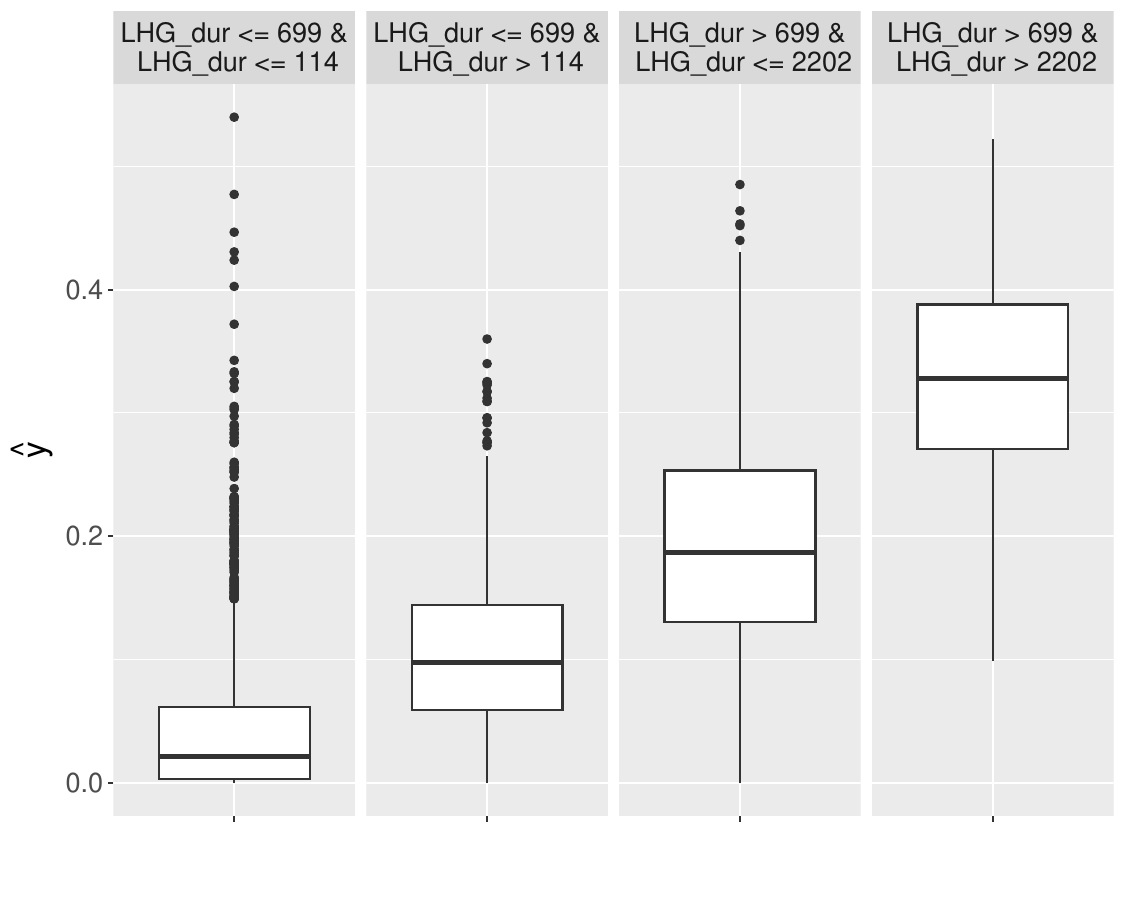}}
\caption{Surrogate model explanations of a random forest predicting long-term unemployment, computed by protected group membership.}
\label{fig:tree}
\end{figure}

In practice, a first step towards merging model interpretation and fairness considerations may include the use of protected attributes as grouping variables to structure the application of IML techniques. Figure \ref{fig:tree}, for example, shows two surrogate decision trees based on the same random forest model which predicts long-term unemployment of job-seekers. In Figure \ref{fig:tree}a, only predictions for German job seekers are used to build the surrogate tree, whereas the tree in Figure \ref{fig:tree}b is based on LTU predictions for non-German job seekers. Note that citizenship was not used as a predictor for the original random forest. In both surrogate trees, the duration of previous unemployment benefit receipt episodes (LHG dur) plays a major role in predicting (future) LTU, with longer receipt histories being associated with higher LTU risk. However, in the surrogate tree for German job seekers older age appears as an additional risk factor. This may indicate that the random forest learned different effect patterns for both subgroups -- a finding which seems reasonable from a performance optimization perspective, but which also points to fairness implications in practice when groups defined by protected attributes are being scored on different grounds. In any case, putting an emphasis on protected groups in the model interpretation process can be helpful to understand if a model may behave differently for important subgroups in downstream applications.

We note some further connections between explainability and fairness. Adversarial attacks are facilitated by an understanding of the model and of the data. Unfortunately, well-intentioned opportunities to probe a model offer such a gateway for attackers. Unchecked access to a model, particularly with IML tools but also under the guise of fairness evaluations, is more threatening than presenting some aggregate evaluation results. 
Of course, attacks can also occur on privacy. Very large models are able to memorize examples (\citealp{belkin.et.al2019.reconciling.modern.ML.classical.bias.variance.tradeoff}; \citealp[ch.~10.8]{james.et.al.2021.intro.statistical.learning}): allowing unchecked access to such models can then have similar consequences as just releasing the original training data would. Members of minority groups $A$, because of their smaller size, may be both, easier to re-identify (because of the small group size) and more likely or vulnerable to suffer negative consequences from re-identification. We are not aware of privacy metrics used as fairness metrics (but see \citealp{wachter.et.al.2017.counterfactual.explanations.without.opening.the.black.box} arguing for privacy-preserving IML).

Individual-level explanations match well with individual-level fairness notions. \textit{Algorithmic recourse} is the notion of giving explanations and recommendations (how to achieve a desired prediction) to the individual, 
particularly in the form of counterfactual explanations \citep{verma.et.al.2020.counterfactual.explanations.algorithmic.recourse.literate.review,karimi.et.al.2021.algorithmic.recourse.survey}. While giving explanations is often desirable, particularly in the context of ADM, the extent to which there exist legal rights to explanations for individuals or legal mandates for organizations gets commonly overestimated (\citealp{doshivelez.et.al.2019.accountability.and.explanations.under.the.law}; \citealp{wachter.et.al.2017.no.right.to.explanation.in.gdpr}). NSOs may also care about individual data points: during data analysis, some may be uncovered as influential to the estimates of a statistical model, and during data processing, some may be flagged as outliers; these might cause performance, robustness, and fairness problems. If such data points were produced by a ML model, e.g., in data imputation, IML-based evaluation of these imputations can help to solve the just-mentioned problems.

\subsection{Cost effectiveness}
In most parts, the costs of adopting ML at NSOs as presented in \cite{yung.et.al.2022.quality.framework.statistical.algorithms} are seen to reflect technical needs such as IT infrastructure, maintenance, and staff training. We want to re-emphasize quality assurance and control as a critical component not only as a means to monitor machine learning models with respect to, e.g., fluctuations in (subgroup) performance, but also as a safety measure: Humans may (need to) overwrite the models' output if the uncertainty exceeds a pre-specified threshold \citep{bhatt2020}. Introducing a `reject option' in supervised learning models, i.e., forwarding difficult cases to humans for classification, can increase error-fairness \citep{kaiser2022}, but by definition comes at the cost of additional manual work. Assessing the need and degree of human oversight thus should be factored into the cost-benefit analysis of high-stakes ML applications at NSOs. Furthermore, fairness cannot be fully automated \citep{weerts.et.al.2023.fairness.auto.ml}.

Some of the new data sources might be cheaper to acquire than traditional data sources, but the savings might not hold up when the additional work to clean up elevated fairness problems is figured in. This is particularly true for `found data' \citep{groves.2011.3.eras.of.survey.research,japec.kreuter.biemer.lane.et.al.2015.big.data.survey.research.aapor.task.force.report} over which NSOs and other stakeholders have little to no discretion in design. Even large sample sizes cannot make up for errors of measurement and representation: if a key variable is subject to differential measurement error or if a protected group is missing it does not matter how many observations are in a data set. A similar argument pertains to data processing: e.g., it might be better to use a medium size survey -- perhaps one that combines survey responses with other data types -- than to use a record linkage model that introduces fairness problems. In data analysis, simpler models have lower demands for data volume and computational resources, for both training and prediction, implying cost and time savings. In addition, higher interpretability may lead to better discovery and removal of fairness problems. Sophisticated, more flexible models might provide more accuracy and thus could be fairer by being more likely to discover model heterogeneity. Thus, both should always be included among the set of models considered for selection.

\subsection{Timeliness}
Adding fairness to the discussion and evaluation of quality dimensions should not be perceived as an additional burden to NSOs. As we try to argue and illustrate throughout this section, fairness considerations can be integrated into existing evaluation procedures in practice and can be viewed as an additional safeguard to ensure that the improvement in timeliness that may be achieved through ML-based automation does not come at the cost of disparate impact downstream. At the same time, the quality dimensions of the QF4SA framework each can benefit from a fairness perspective as it enriches the evaluation of algorithms by highlighting the critical role of (social) subgroups.

An interesting development that implicitly links fairness to timeliness is the work on fairness-aware automated machine learning \citep{weerts2023}. In this line of research, methods are being proposed that can improve timeliness and cost by automating parts of the machine learning pipeline, while the resulting output is also required to fulfill some fairness constraints. While it is important to recognize the limits of such an approach -- the authors agree that fairness cannot be fully automated --, fairness-aware AutoML can still expand the methodological toolkit of NSOs.

Timeliness, cost of data collection, and overall sample size are reasons to try to predict, e.g., response propensity in surveys. It has been recognized that focusing recruitment efforts on units with a high predicted propensity to participate is tempting on the aforementioned dimensions but widens the potential for representation bias when response propensity also depends on the outcome variable $Y$ for the eventual data analysis. From a fairness perspective, it is important to note that hard-to-survey or hard-to-reach subpopulations \citep{tourangeau.et.al.2014.hard.to.survey.populations.edited.volume,willis.et.al.2014.overview.special.issue.hard.to.reach} may often coincide with groups for which we worry about discrimination and biased outputs. Note that even if there is no relation between outcome and response propensity in a group, if the group is small in the collected data, the power to detect model heterogeneity is diminished and fairness evaluations become more statistically uncertain.

\subsection{Robustness}
The lack of robustness and stability can be connected to fairness concerns in multiple ways. On the organizational level, model decay or drift (i.e., deteriorating model accuracy over time) can be a reason for (potentially selective) skepticism towards algorithmic solutions \citep[p.~2]{ml.group.2022.model.retraining}. In downstream applications, model drift can affect different parts of the target population in different ways. That is, differential error across subgroups may surface or may be amplified due to shifts in the data to which the model is applied. It may also be harder to detect model drift that occurs mainly or first in (small) protected groups. Also, one type of drift or drift indicator, namely the emergence of new categories in a categorical feature or outcome variable, might itself be directly about the existence and recognition of protected groups. 
To our knowledge, monitoring of models and decisions about the need to re-train to date consider global performance measures (e.g., \citealt{ml.group.2022.model.retraining}). From the fairness perspective, we suggest that (sub)group measures must also be monitored: this will not only inform when error-fairness drops below a pre-specified threshold, but might also inform about the causes and possible countermeasures. We further argue that careful monitoring is also needed even if models are re-trained on a regular basis, as new biases may be picked up along the way.

Furthermore, the robustness of a model within a group and the (epistemic) uncertainty of a model's predictions for a group have, to our knowledge, not been seen as fairness criteria so far. We suggest that these desirable global model properties should be also investigated as individual or (sub)group fairness notions and metrics in the algorithmic fairness literature. This pertains to models that are used in data collection, data processing, and data analysis.

\phantomsection\label{robust.fairness.causal}
One way to robustify models against drifts 
\citep[ch.~9.4.3]{varshney.2022.trustworthy.ml.book} is to employ causal models: e.g., causal relationships rooted in physics or biology are assumed to be more stable over time than spurious correlations.\footnote{
    It is also plausible that causal features are harder to game than spurious features (which have no effect on $Y$ and thus may be changed at low `cost'), making causal models more robust to adversarial attacks.
} 
Causal features and causal fairness notions are also being discussed (e.g, \citealp{makhlouf.et.al.2022.survey.causal.fair.ml,plecko.barenboim.2022.causal.fairness.analysis}). 
Yet, while employing causal features may be attractive in some settings,  for images or other unstructured data types analyzed with Deep Learning, traditional features (in a potentially causal sense) are hardly involved.

\begin{figure}[h!]
\centering
\includegraphics[scale = 0.55]{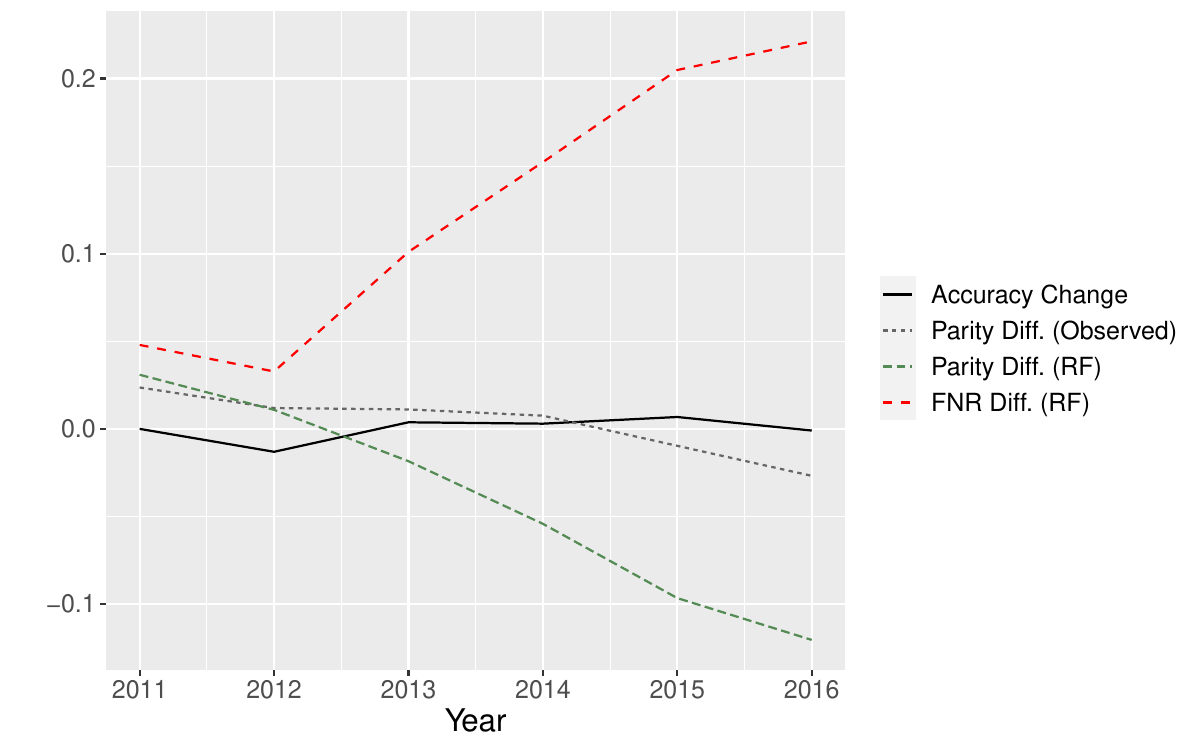}
\caption{(Change in) prediction performance and selected fairness metrics for random forest models over time. For each year, a new random forest is trained and evaluated with data from the next year. Parity difference scores show the difference in predicted LTU rates between non-German and German job seekers. FNR difference scores show the difference in false negative rates between non-Germans and Germans.}
\label{fig:drift}
\end{figure}

Monitoring fairness metrics can be particularly important in a deployment context that includes data sources that capture complex, natural processes. In Figure \ref{fig:drift} we hold the model design constant, that is, random forest models for predicting long-term unemployment are used with the same hyper-parameter settings and features, but we repeatedly train and test models with data that change over time. Specifically, we use labor market records from 2010--2016 and train one random forest model for each year, and evaluate the respective model with data from the next year. The bold black line shows the difference in overall model performance (balanced accuracy) as we move from one year to the next. From this point of view, we might conclude that we can safely apply our random forest modeling schema over time without any major disruptions. However, assessing fairness metrics points to a different conclusion: a considerable increase in false negative rate (FNR) differences between non-German and German job seekers (dashed red line) can be observed when training and evaluating models with more recent data. This is accompanied by increasing parity differences in the models' predictions (dashed green line), which over-amplify the true differences in base rates as observed in the data (dotted gray line). As such changes over time can have considerable implications in practice, requiring robustness assessments to also consider subgroup-specific (fairness) metrics appears advisable.  

\subsection{Reproducibility}
From a fairness perspective, (inferential) reproducibility raises questions as to how strongly design decisions in the machine learning pipeline affect outcomes not just overall, but also separately for sensitive subgroups of the target population. Fairness-relevant decision points may not only include the machine learning model itself (e.g., the model type and hyperparameter settings), but also more subtle aspects that include implicit decisions in data pre-processing steps (e.g., NSOs may employ a standard procedure for imputing missing values, while different imputation strategies can affect fairness measures in different ways; \citealt{caton2022}). In practice, the implications of non-reproducibility may again be assessed by structuring model evaluations by protected attributes, paired with a grid of design decisions that is centered around the intended deployment setup.

\begin{figure}[h!]
\centering
\subfloat[Similarities between LTU predictions for Germans]{\includegraphics[scale = 0.4]{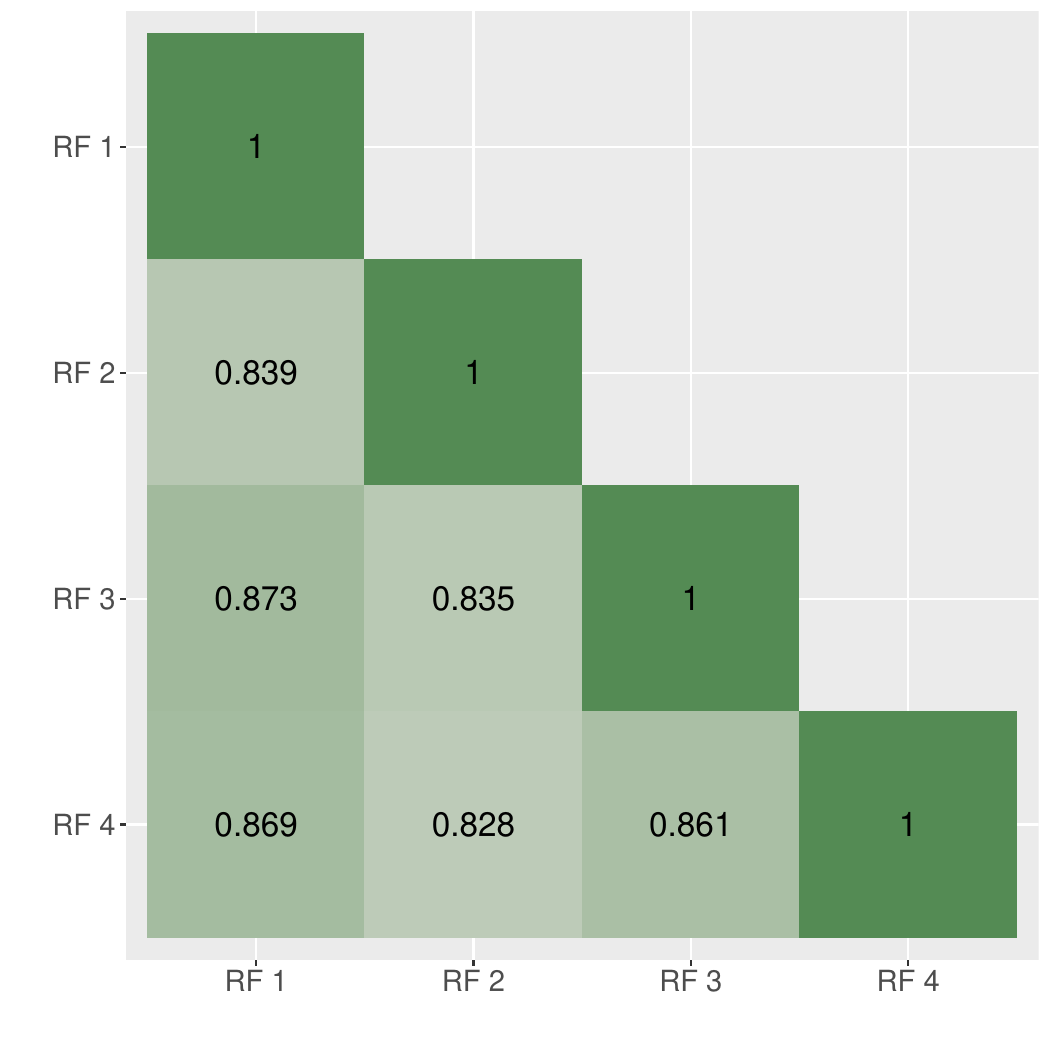}}%
\quad
\subfloat[Similarities between LTU predictions for non-Germans]{\includegraphics[scale = 0.4]{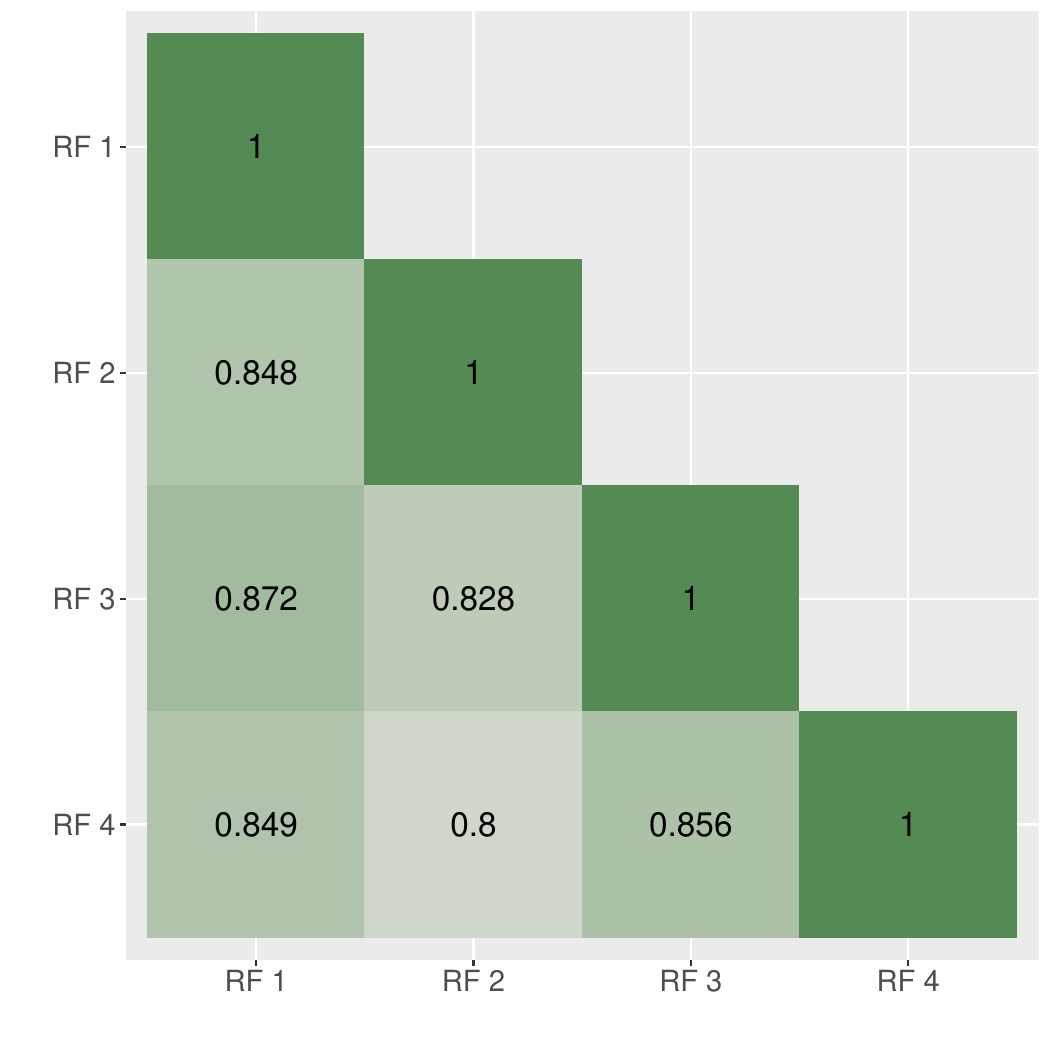}}
\caption{Jaccard similarities between LTU predictions of random forest models with different hyper-parameter settings (RF 1: ntree = 750, nodesize = 1, RF 2: ntree = 250, nodesize = 1, RF 3: ntree = 500, nodesize = 5, RF 4: ntree = 500, nodesize = 15), computed by protected group membership.}
\label{fig:similarity}
\end{figure}

A strong susceptibility to design decisions is of particular concern if the model outputs are further used downstream, either as an input to further analysis or to directly inform actions. Figure \ref{fig:similarity} focuses on the effects of different hyperparameter settings on the classifications of random forest models predicting long-term unemployment. Four forests were trained that differ in the number of trees and the minimum size of the trees' terminal nodes and are then used to predict LTU, using the same classification threshold (top 25\%). The Jaccard similarities, denoting the overlap (between 0 and 1) between the LTU predictions of the different random forest models are plotted, separately for German (Figure \ref{fig:similarity}a) and non-German (Figure \ref{fig:similarity}b) job seekers. Considering the modest changes that were made in the random forests' setup, we observe non-trivial differences between the lists of job seekers that are predicted as being at high risk of LTU by each model. While this generally holds for both German and non-German job seekers, the lowest agreement in predictions is recorded in Figure \ref{fig:similarity}b (between RF 2 and 4). Assessing the susceptibility of outcomes to small changes in the modeling design with a focus on protected groups thus may allow to identify variation that can challenge both overall reproducibility and the consistency of outcomes for societal subgroups.

As stressed in section \ref{subsec:reproduc}, methods reproducibility is increasingly viewed as a minimum standard. If the root causes of a model's discovered fairness problems are investigated upstream, the respective models that were used, e.g., in data processing must be reproducible or the search for problems and solutions may be futile.

\subsection{Accuracy}
Fairness interacts with accuracy (and with the human component) at multiple stages of the production process of NSOs. In the context of data processing and preparation, we note that one of the purported benefits of automation is an increase in consistency: e.g., in annotation tasks, even subject matter experts can disagree \citep[p.~12]{sthamer.2020.classification.coding.theme.report} and, over time, an annotator might become tired or less motivated (inter- and intra-annotator reliability, respectively).\footnote{
    The same is true for editing and imputation (\citealp[p.~6]{dumpert.2020.unece.editing.imputation.theme.report}; \citealp{sthamer.2020.editing.social.survey.data.with.ML}) and other data processing tasks. We will focus on annotation as an example.
} 
Meanwhile, a model will `decide' the same way every time -- but it is trained to reproduce the patterns contained in the training data, including those made by human annotators. 
First, consider a single annotator. She or he is or feels required to provide a label even for difficult-to-decide cases. Absent any other option, the annotator might resort to the marginal distribution of $Y|A$ (or their subjective notion thereof), even when $Y$ is independent of $A$ given $X$. The trained model will then learn that $A$ (or proxies of $A$) are predictive of the provided labels (even though it is not predictive of the true $Y$). Solutions may include letting annotators express uncertainties instead of forcing a choice; this may actually fit well with the aforementioned reject-option models. 
Also, information about $A$ may be better hidden from annotators, although it can be difficult to do so in, e.g., image-based classification tasks. 
A second mechanism of how annotations can induce biased models predictions is a non-random allocation of observations to a set of inherently heterogeneous annotators: e.g., if units from $A=a$ are mostly processed by an annotator with a low general propensity to label $Y=1$ and, conversely, units from $A \neq a$ are mostly processed by an annotator with a high general propensity for $Y=1$, again $A$ becomes predictive of the labels even though when it is not related to the true values $Y$. This is an example of how biases can be introduced during data processing even when there is no overt discrimination. A similar argument pertains to data collection, e.g., the allocation of interviewers might cause measurement or representation errors. 
We see two options here: NSOs can use stratified randomization to allocate data points to annotators and they can release annotator IDs because, conditional on the annotator, the spurious relationship of $A$ and the labels vanishes.

Aggregation of data is one of the core tasks of NSOs: both in terms of data analysis, which may be simple descriptive statistics, and in terms of producing (aggregate) data for release. 
For the former, ML may seem hardly helpful for the estimation of population parameters: Other than (short) trees, ML model classes hardly possess parameters that correspond to interpretable, meaningful population characteristics. Also, systematically biased estimation, due to the bias-variance trade-off caused by training supervised ML models to minimize the expected prediction error, is problematic. However, ML can be useful when a parameter of interest is not identical across all subpopulations. For up to a medium number of pre-identified subpopulations, multiple testing correction can be employed to limit the error of falsely claiming heterogeneity \citep[ch.~4.1]{athey.lab.2023.ml.causal.inference.tutorial}. If there are, however, many subpopulations to investigate, as is the case with intersectional fairness, or if there are no pre-specified hypotheses at all, ML can help to discover heterogeneity: data splitting is then the procedure that guards against false discovery \citep[ch.~4.2]{athey.lab.2023.ml.causal.inference.tutorial}. If interpretable heterogeneity is the goal, trees for univariate statistics or, for more complex analyses, causal trees \citep{athey.imbens.2016.causal.trees} appear to be most suitable.\footnote{
    Such methods may be developed mostly for causal inference, but (finding) heterogeneity is also relevant for more descriptive data analysis. 
} 
We suggest that NSOs use such methodology for more fair reporting of the results of data analysis: subgroups for which the parameter deviates more than a pre-specified, meaningful amount from the global average should be identified and reported along with the global average. 
Aside from fairness, this approach can also be used to determine whether the global parameter value is meaningful and worth reporting at all: from Simpson's paradox, it is well known that the global value may be completely different than the value in all subpopulations, e.g., taking on a different sign.

For error-based fairness notions, the same methodology can be used to find subpopulations that suffer from more errors. Similarly, if there is gold-standard evaluation data for a data production process whether based on human work, ML, traditional statistics, or a combination thereof, such as editing and imputation, NSOs can find groups for which their process performs more poorly compared to others or to an absolute threshold. If such heterogeneity is found, it may help -- but will not replace subject matter and data knowledge -- in fixing deficiencies in data collection (e.g., improving survey questions to yield less measurement error for $A$) and data processing systems. 
To our knowledge, using heterogeneity-finding ML machinery has not been explicitly suggested in the fair ML literature for either fair reporting of data analysis or finding biases in the data (production process), potentially with the exception of \cite{vonzahn2023locating}.

Self-assessed confidence measures by ML models may be used to decide whether something should be labeled by the model or be referred to a human expert (e.g., \citealt{unece.2022.text.classification.theme.group.report} for text classification). However, not every ML model class yields self-assessed uncertainty measures and for those that do, there is no guarantee that they are accurate on average. Moreover, a model's overconfidence may not be the same for every group but could be worse for some (protected) groups or individuals. Some uncertainty measures also only recognize aleatoric, but not epistemic uncertainty and the latter may be greater for small minorities $A$. Uncertainty evaluations thus must also use actual evaluation data and cannot solely rely on models' self-assessments.

\begin{figure}[h!]
\centering
\includegraphics[scale = 0.475]{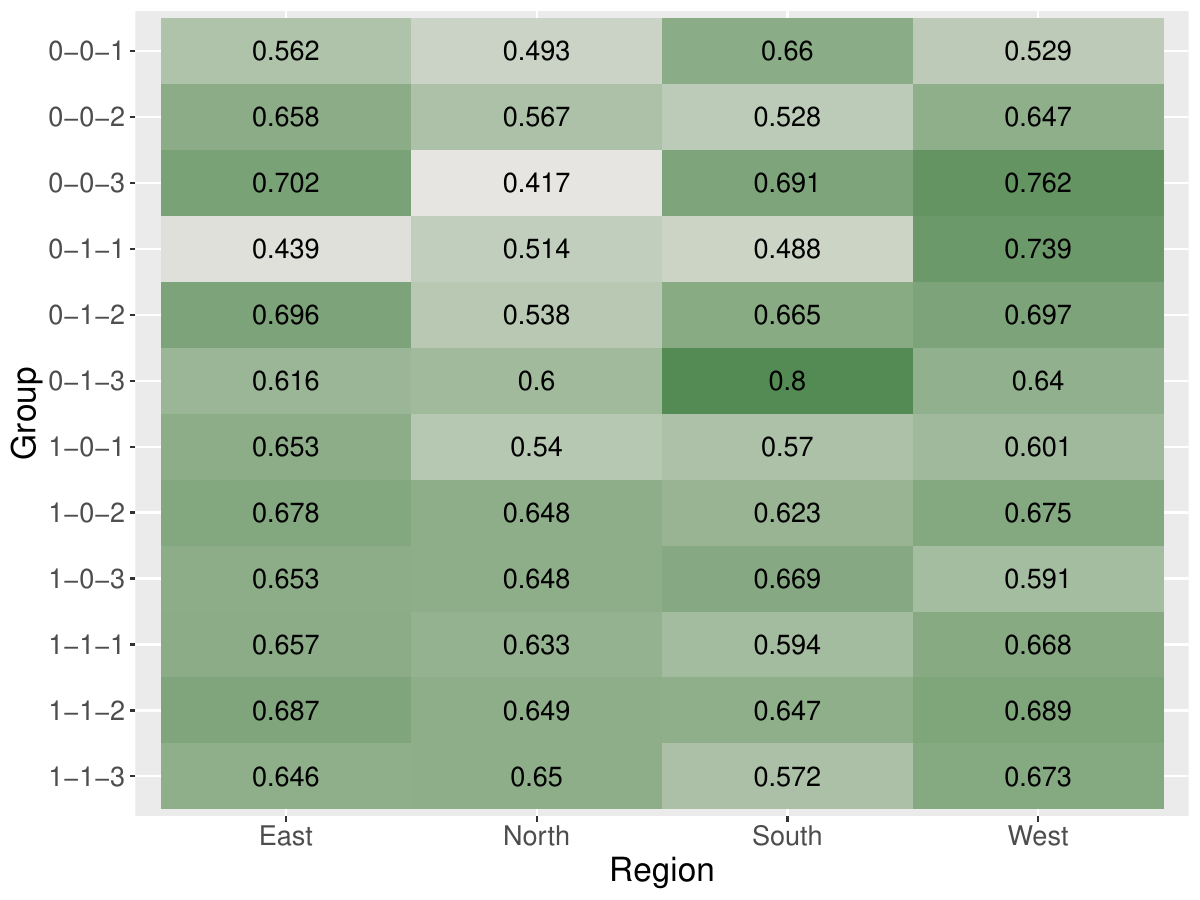}
\caption{Subgroup prediction performance (balanced accuracy) of a random forest predicting long-term unemployment. Group coding scheme: Citizenship (0: non-German, 1: German) -- Gender (0: Male, 1: Female) -- Age group (1: 18--30, 2: 31--50, 3: $>$50).}
\label{fig:subgroup}
\end{figure}

In a supervised learning context, accuracy as a quality dimension can be naturally extended to capture fairness concerns by requiring accurate predictions not just overall, but also for subgroups which may be defined by protected attributes or other features that are viewed as substantively relevant in a given application \citep{kim_multiaccuracy_2019, hebert-johnson_multicalibration_2018}. Based on our long-term unemployment prediction example, Figure \ref{fig:subgroup} shows balanced accuracy scores of a random forest predicting LTU computed for 48 subgroups in the test set. Specifically, subgroups of job seekers were defined by intersections of the attributes citizenship, gender, age group, and region. While the model achieves an overall balanced accuracy score of 0.667, considerable variation in subgroup performance can be observed. Accuracy ranges between 0.417 and 0.8, indicating that prediction performance is no better, and sometimes worse, than random guessing for some demographic subgroups. The strongest variation in scores can be observed for non-German job seekers (upper half of Figure \ref{fig:subgroup}), i.e., for subsets of the minority group in our example. While this finding may in part be driven by small sample sizes in some cells (although all but three cells include more than 50 observations), it highlights the utility of assessing subgroup accuracy as a means to provide pointers for further model investigation.    

%% file: fairness.other.tex
One mechanism causing fairness problems is what we call unrecognized model heterogeneity: i.e., the true functional relationship between features and outcome for units from some group $A$ is not identical to the relationship in the rest of the population. If there are too few examples from $A$ in the training data, the power to detect the correct model for $A$ is low. There are several upstream causes for this phenomenon. Coverage errors, sampling errors, or unit nonresponse patterns may be such that members of $A$ are underrepresented. 
Processing may also contribute. Unsupervised outlier detection methods identify unusual data points: thus, units from a small minority $A$ are at risk for being falsely detected and removed -- not because of erroneous values, but simply because of their membership in an infrequent group. 

Unsupervised identification for record linkage or duplicate removal can also be sensitive to group membership. For instance, name-based distance metrics may be impacted when foreign names have multiple transliterations into the NSO's language or when, e.g., self-chosen Western first names are used in one data set and the original, non-Western first name in the other. Also, the relative frequency of first and last name combinations may be higher or lower for members of some group $A$ than in the general population, hurting or benefiting their record linkage success. Supervised record linkage may be trained on data sources that contained relatively few recognized links (i.e., the label in this case) for members of $A$. This could be because of differential label error. It could also be that a low base rate (of correct links for members of $A$) was correct for the original training data sources, but are not for the data sources one currently wants to link.

NSOs as data producers can investigate at which processing steps many units from group $A$ were lost, in relative or absolute terms
. Two questions arise. First, which groups should be considered? A general, standard canon of groups to consider plus application-specific groups based on subject matter knowledge are obvious starting points, and is the focus of algorithmic fairness literature shaped by the notion of protected attributes defined by law. This is also the subject of ethical and legal discussions that methodologists can and should not resolve on their own. 
We suggest supplementing this with a data-driven approach: groups for which the loss of units, in absolute terms or relative to the rest, is above a certain threshold. The loss of units can be calculated with regard to the previous step in multi-step data processing or, where applicable, with respect to the true distribution of characteristics in the target population (based on large-scale, gold-standard distributional data such as censuses). Second, which criteria should be applied? How to set the threshold for permissible relative loss of units in order to improve internal processes and the data is an organizational decision, based on the available resources. The criterion for absolute loss of units should be tied to the consequence, the loss of statistical power. How many units are needed to detect model heterogeneity beyond a certain level for $A$? How many units are needed for fairness evaluations with a pre-specified tolerable uncertainty? 
Unfortunately, the answers are largely application-specific and NSOs cannot anticipate all possible applications by external data users. Still, fairness report cards and metadata for released data should include information on losses of units that exceed thresholds. Users should be put in the position to be able to decide whether a certain product fits their needs and fairness demands -- outright, after supplementation with other data sources, or not at all.

%% file: conclusion.tex
The advent of \textit{automated} decision-making and the rigorous focus on performance inherent to the ML mindset likely both contributed to the rise of fair ML. We argued in this paper that fairness is also a desirable, perhaps even necessary quality dimension of the work of NSOs -- similar to how fairness is one dimension of frameworks for Trustworthy ML. This is, more generally, true for all data collection, processing, or analysis processes in official statistics: those that use ML or automation, but also those that employ traditional methods or human work. Nonetheless, the deployment of ML re-amplifies the need for explainable, reproducible, robust, and accurate products and data production processes at NSOs, highlighting quality dimensions that critically interact with fairness considerations as outlined in this article.

We further discussed the importance of the \textit{human component} (section \ref{subsec:human}) in (fair) ML at NSOs.
In the pure ML world, some may believe that domain knowledge is unnecessary and that ML models, enough data, and ML knowledge are all that is required. The `end-to-end' promise of Deep Learning being able to turn (seemingly) raw data into the desired predictions may add to that view.\footnote{
    We caution that many applications, particularly those working with survey data, lack the enormous training data required to render (knowledge-based) feature engineering obsolete.} 
We believe it is unwarranted. For instance, for some unlabeled data, e.g., images, subject matter expertise is required to produce the high-quality annotations on which the model's eventual success depends \citep[p.~2]{julien.2020.unece.hlg.mos.ml.project.report}. For models that assist staff in, e.g., coding, what makes good suggestions and how humans interact with the model's output can be highly context-dependent.

It is similar with fairness. Our suggestion of data-driven finding and reporting of unfairness (e.g., section \ref{subsec:accuracy}) is a complement, not a replacement for legal knowledge and ethical considerations. For instance, debating which groups might be impacted the most and thus deserve fairness evaluations requires knowledge of the specific context and the general working of society. Once fairness problems have been detected, the work to find the causes and solutions starts. This is especially true for NSOs for whom not publishing data or statistics that are too unfair is often not an option, 
but who instead must find a way to improve. Finding discrimination in the data or making the annotation process less biased are among the tasks that require e.g., subject matter experts, statisticians, and methodologists. 

Another aspect of the importance of the human factor is the willingness to accept systems that involve ML. Beyond the macro level, the different individual stakeholders need to be on board \citep[ch.~6f.]{julien.2020.unece.hlg.mos.ml.project.report}: e.g., from the (internal and external) users of a system and its output to anyone whose work is affected, such as experts whose roles are shifted. There are two core factors \citep[ch.~6]{julien.2020.unece.hlg.mos.ml.project.report}: First, such systems must demonstratedly serve the individual and organizational ``business needs''. Second, a trusted quality framework must form the basis: it guides the workflow (to prevent problems) and the actual performance on its quality dimensions is transparently and credibly evaluated. Fairness, as its own quality dimension, in its interaction with the other dimensions, and as part of legal and ethical considerations plays a big part in this. 

Lastly, even if individuals are hesitant to embrace new ML methods outright, it can be still advisable to broaden the toolbox: if in comparison a statistical method performs similarly well, one can, in good conscience, use the more known, and interpretable traditional method; if the statistical method is however vastly outperformed, then this is at least a call to critically assess violations of the assumptions baked into the statistical method. At any rate, there is no need to let these disadvantages keep institutions from profiting from the positives of ML methods.

We further emphasize the critical role of data quality and its interaction with (fair) ML at NSOs. It is no secret that ML applications depend on the quantity and, although sometimes neglected, on the quality of training data. Relative to other data producers, NSOs have a long track record, extensive expertise, and legal obligation to (data) quality principles (e.g., \citealp[p.~7]{eurostat.2017.eu.code.of.practice}; \citealp[ch.~2]{julien.2020.unece.hlg.mos.ml.project.report}). We believe that NSOs as a whole also have a competitive advantage because of their commitment to collaboration (\citealp[Principles~8-10]{unece.2013.fundamental.principles.official.statistics}; \citealp[Principle~1bis]{eurostat.2017.eu.code.of.practice}): beyond the sharing of code and knowledge \citep[ch.~7]{julien.2020.unece.hlg.mos.ml.project.report}, we suggest the different entities can pool training data and share in the expensive, but crucial human annotation tasks. This will increase efficiency and cross-organizational consistency.
While the fairness dimension implies further requirements for the metadata and other documentation to be released alongside with NSOs' data and statistics products, 
the transparent publishing of such valuable, credible documentation can also be seen as a competitive advantage of NSOs over their competitors \citep[ch.~2]{julien.2020.unece.hlg.mos.ml.project.report} and of NSOs' products over `found data'. We thus argue that fairness need not be seen as an additional burden, but rather caters toward the key objective of NSOs of releasing high-quality data products.  

%% file: abbrev.tex
\begin{table}[!ht]
\begin{tabular}{ll}	
    \hline
	AI		& 	Artificial Intelligence	\\
	DGP		&	Data-generating process\\
    ESPE    &   Expected squared out-of-sample prediction error\\
    FNR     &   False Negative Rate \\
	IML		&	Interpretable Machine Learning	\\
    LLM     &   Large Language Model\\
    LTU     &   Long-term unemployment\\
	ML		&	Machine Learning	\\
	NSO		&	National Statistical Organization	\\
    TSE      &   Total Survey Error\\
	QF4SA	&	\cite{yung.et.al.2022.quality.framework.statistical.algorithms}'s Quality Framework for Statistical Algorithms	\\
	XAI		&	Explainable Artificial Intelligence	\\
	\hline
\end{tabular}
\caption{List of abbreviations}
\label{tab:tab.abbrev}
\end{table}



%% file: main.bbl
\begin{thebibliography}{}

\bibitem[{AlgorithmWatch}, 2019]{algorithm.watch.2019.atlas.of.automation}
{AlgorithmWatch} (2019).
\newblock {Atlas of Automation. Automated decision-making and participation in
  Germany}.

\bibitem[Allhutter et~al., 2020]{allhutter_algorithmic_2020}
Allhutter, D., Cech, F., Fischer, F., Grill, G., and Mager, A. (2020).
\newblock Algorithmic {Profiling} of {Job} {Seekers} in {Austria}: {How}
  {Austerity} {Politics} {Are} {Made} {Effective}.
\newblock {\em Frontiers in Big Data}, 3:5.

\bibitem[Amaya et~al.,
  2020]{amaya.biemer.kinyon.2020.total.error.framework.big.data.from.tse}
Amaya, A., Biemer, P.~P., and Kinyon, D. (2020).
\newblock {{Total Error in a Big Data World: Adapting the TSE Framework to Big
  Data}}.
\newblock {\em Journal of Survey Statistics and Methodology}, 8(1):89--119.

\bibitem[Angelopoulos and Bates,
  2022]{angelopoulos.bates.2022.gentle.intro.to.conformal.prediction}
Angelopoulos, A.~N. and Bates, S. (2022).
\newblock {A Gentle Introduction to Conformal Prediction and Distribution-Free
  Uncertainty Quantification}.

\bibitem[Angwin et~al., 2016]{angwin_machine_2016}
Angwin, J., Mattu, S., and Kirchner, L. (2016).
\newblock Machine {Bias}.
\newblock Technical report, ProPublica.

\bibitem[Antoni et~al., 2019]{siab2019}
Antoni, M., Ganzer, A., and vom Berge, P. (2019).
\newblock {Sample of Integrated Labour Market Biographies Regional File
  (SIAB-R) 1975 - 2017}.
\newblock FDZ-Datenreport, 04/2019 (en).
\newblock https://doi.org/10.5164/IAB.FDZD.1904.en.v1.

\bibitem[Argyle et~al.,
  2023]{argyle.et.al.2023.gpt.to.simulate.human.respondents}
Argyle, L.~P., Busby, E.~C., Fulda, N., Gubler, J.~R., Rytting, C., and
  Wingate, D. (2023).
\newblock {Out of One, Many: Using Language Models to Simulate Human Samples}.
\newblock {\em Political Analysis}, pages 1--15.
\newblock published online 2023/02/21.

\bibitem[Athey and Imbens, 2016]{athey.imbens.2016.causal.trees}
Athey, S. and Imbens, G. (2016).
\newblock Recursive partitioning for heterogeneous causal effects.
\newblock {\em Proceedings of the National Academy of Sciences},
  113(27):7353--7360.

\bibitem[Baker,
  2017]{baker.2017.big.data.survey.research.perspective.in.tse.book}
Baker, R. (2017).
\newblock {Big Data: A Survey Research Perspective}.
\newblock In Biemer, P.~P., de~Leeuw, E.~D., Eckman, S., Edwards, B., Kreuter,
  F., Lyberg, L.~E., Tucker, N.~C., and West, B.~T., editors, {\em {Total
  Survey Error in Practice}}, pages 47--69. John Wiley \& Sons, Hoboken, NJ.

\bibitem[Barba, 2018]{barba.2018.terminologies.for.reproducible.research}
Barba, L.~A. (2018).
\newblock {Terminologies for Reproducible Research}.

\bibitem[Barocas et~al., 2019]{barocas_fairness_2019}
Barocas, S., Hardt, M., and Narayanan, A. (2019).
\newblock {\em Fairness and {Machine} {Learning}}.
\newblock fairmlbook.org.

\bibitem[Barocas and Selbst, 2016]{barocas_big_2016}
Barocas, S. and Selbst, A.~D. (2016).
\newblock Big {Data}'s {Disparate} {Impact}.
\newblock {\em California Law Review}, 104(3):671--732.

\bibitem[Beck et~al.,
  2022]{beck.et.al.2022.improving.labeling.through.social.science.research.agenda}
Beck, J., Eckman, S., Chew, R., and Kreuter, F. (2022).
\newblock {Improving Labeling Through Social Science Insights: Results and
  Research Agenda}.
\newblock In Chen, J. Y.~C., Fragomeni, G., Degen, H., and Ntoa, S., editors,
  {\em HCI International 2022 -- Late Breaking Papers: Interacting with
  eXtended Reality and Artificial Intelligence}, pages 245--261, Cham,
  Switzerland. Springer.

\bibitem[Beck et~al.,
  2018a]{beck.dumpert.feuerhake.2018.ml.in.official.statistics}
Beck, M., Dumpert, F., and Feuerhake, J. (2018a).
\newblock {Machine Learning in Official Statistics}.
\newblock {arXiv}, \url{https://arxiv.org/abs/1812.10422}.

\bibitem[Beck et~al.,
  2018b]{beck.dumpert.feuerhake.2018.proof.of.concept.ml.abschlussbericht}
Beck, M., Dumpert, F., and Feuerhake, J. (2018b).
\newblock {Proof of Concept Machine Learning}.
\newblock Abschlussbericht, Federal Statistical Office of Germany, Wiesbaden,
  Germany.

\bibitem[Belkin et~al.,
  2019]{belkin.et.al2019.reconciling.modern.ML.classical.bias.variance.tradeoff}
Belkin, M., Hsu, D., Ma, S., and Mandal, S. (2019).
\newblock Reconciling modern machine-learning practice and the classical
  bias–variance trade-off.
\newblock {\em Proceedings of the National Academy of Sciences},
  116(32):15849--15854.

\bibitem[Bellamy et~al., 2022]{bellamy.hernan.beam.2022.shortcut.features}
Bellamy, D., Hern{\'{a}}n, M.~A., and Beam, A. (2022).
\newblock A structural characterization of shortcut features for prediction.
\newblock {\em European Journal of Epidemiology}, 37(6):563--568.

\bibitem[Bengs et~al.,
  2022]{bengs.huellermeier.waegeman.2022.difficulty.of.epistemic.uncertainty.quantification}
Bengs, V., H{\"{u}}llermeier, E., and Waegeman, W. (2022).
\newblock {On the Difficulty of Epistemic Uncertainty Quantification in Machine
  Learning: The Case of Direct Uncertainty Estimation through Loss
  Minimisation}.

\bibitem[Bhatt et~al., 2020]{bhatt2020}
Bhatt, U., AntorÃ¡n, J., Zhang, Y., Liao, Q.~V., Sattigeri, P., Fogliato, R.,
  MelanÃ§on, G.~G., Krishnan, R., Stanley, J., Tickoo, O., Nachman, L.,
  Chunara, R., Srikumar, M., Weller, A., and Xiang, A. (2020).
\newblock Uncertainty as a form of transparency: Measuring, communicating, and
  using uncertainty.

\bibitem[Binns, 2018]{binns_fairness_2018}
Binns, R. (2018).
\newblock Fairness in {Machine} {Learning}: {Lessons} from {Political}
  {Philosophy}.
\newblock {\em arXiv:1712.03586 [cs]}.

\bibitem[Bommasani et~al.,
  2021]{bommasani.liang.et.al.2021.opportunities.risks.foundation.models}
Bommasani, R., Hudson, D.~A., Adeli, E., Altman, R., Arora, S., von Arx, S.,
  Bernstein, M.~S., Bohg, J., Bosselut, A., Brunskill, E., Brynjolfsson, E.,
  Buch, S., Card, D., Castellon, R., Chatterji, N., Chen, A., Creel, K., Davis,
  J.~Q., Demszky, D., Donahue, C., Doumbouya, M., Durmus, E., Ermon, S.,
  Etchemendy, J., Ethayarajh, K., Fei-Fei, L., Finn, C., Gale, T., Gillespie,
  L., Goel, K., Goodman, N., Grossman, S., Guha, N., Hashimoto, T., Henderson,
  P., Hewitt, J., Ho, D.~E., Hong, J., Hsu, K., Huang, J., Icard, T., Jain, S.,
  Jurafsky, D., Kalluri, P., Karamcheti, S., Keeling, G., Khani, F., Khattab,
  O., Koh, P.~W., Krass, M., Krishna, R., Kuditipudi, R., Kumar, A., Ladhak,
  F., Lee, M., Lee, T., Leskovec, J., Levent, I., Li, X.~L., Li, X., Ma, T.,
  Malik, A., Manning, C.~D., Mirchandani, S., Mitchell, E., Munyikwa, Z., Nair,
  S., Narayan, A., Narayanan, D., Newman, B., Nie, A., Niebles, J.~C.,
  Nilforoshan, H., Nyarko, J., Ogut, G., Orr, L., Papadimitriou, I., Park,
  J.~S., Piech, C., Portelance, E., Potts, C., Raghunathan, A., Reich, R., Ren,
  H., Rong, F., Roohani, Y., Ruiz, C., Ryan, J., Ré, C., Sadigh, D., Sagawa,
  S., Santhanam, K., Shih, A., Srinivasan, K., Tamkin, A., Taori, R., Thomas,
  A.~W., Tramèr, F., Wang, R.~E., Wang, W., Wu, B., Wu, J., Wu, Y., Xie,
  S.~M., Yasunaga, M., You, J., Zaharia, M., Zhang, M., Zhang, T., Zhang, X.,
  Zhang, Y., Zheng, L., Zhou, K., and Liang, P. (2021).
\newblock {On the Opportunities and Risks of Foundation Models}.
\newblock v3, 12 Jul 2022.

\bibitem[Bothmann et~al., 2022]{Bothmann2022}
Bothmann, L., Peters, K., and Bischl, B. (2022).
\newblock What is fairness? {Implications} for {FairML}.
\newblock \url{https://arxiv.org/abs/2205.09622}.

\bibitem[Bothmann et~al.,
  2023]{bothmann.et.al.2023.automated.wildlife.image.classification}
Bothmann, L., Wimmer, L., Charrakh, O., Weber, T., Edelhoff, H., Peters, W.,
  Nguyen, H., Benjamin, C., and Menzel, A. (2023).
\newblock {Automated wildlife image classification: An active learning tool for
  ecological applications}.

\bibitem[Buolamwini and Gebru, 2018]{Buolamwini2018}
Buolamwini, J. and Gebru, T. (2018).
\newblock Gender shades: Intersectional accuracy disparities in commercial
  gender classification.
\newblock In Friedler, S.~A. and Wilson, C., editors, {\em Proceedings of the
  1st Conference on Fairness, Accountability and Transparency}, volume~81 of
  {\em Proceedings of Machine Learning Research}, pages 77--91. PMLR.

\bibitem[Burton et~al.,
  2020]{burton.et.al.2020.algorithm.aversion.literature.review}
Burton, J.~W., Stein, M.-K., and Jensen, T.~B. (2020).
\newblock A systematic review of algorithm aversion in augmented decision
  making.
\newblock {\em Journal of Behavioral Decision Making}, 33(2):220--239.

\bibitem[Carroll et~al., 2006]{carroll.et.al.2006.measurement.error}
Carroll, R.~J., Ruppert, D., Stefanski, L.~A., and Crainiceanu, C.~M. (2006).
\newblock {\em {Measurement Error in Nonlinear Models: A Modern Perspective}}.
\newblock Chapman and Hall/CRC, Boca Raton, FL, 2nd edition edition.

\bibitem[Caton and Haas, 2020]{Caton2020}
Caton, S. and Haas, C. (2020).
\newblock Fairness in machine learning: A survey.

\bibitem[Caton et~al., 2022]{caton2022}
Caton, S., Malisetty, S., and Haas, C. (2022).
\newblock Impact of imputation strategies on fairness in machine learning.
\newblock {\em Journal of Artificial Intelligence Research}, 74.

\bibitem[Chen et~al.,
  2019]{chen.et.al.2019.ai.in.medicine.potential.and.shortcut.learning}
Chen, J., Beam, A., Saria, S., and Mendon{\c{c}}a, E.~A. (2019).
\newblock {Potential Trade-Offs and Unintended Consequences of Artificial
  Intelligence}.
\newblock In Matheny, M., Israni, S.~T., Ahmed, M., and Whicher, D., editors,
  {\em {Artificial Intelligence in Health Care: The Hope, the Hype, the
  Promise, the Peril}}, page~89. National Academy of Medicine, Washington, DC.

\bibitem[Choi et~al., 2022]{ml.group.2022.model.retraining}
Choi, I., del Monaco, A., Law, E., Davies, S., Karanka, J., Baily, A., Piela,
  R., Turpeinen, T., Mharzi, A., Rastan, S., Flak, K., and Jentoft, S. (2022).
\newblock {ML Model Monitoring and Re-training in Statistical Organisations}.
\newblock ONS-UNECE Machine Learning Group 2022, Theme Group - Model
  Retraining, v2, available at
  \url{https://statswiki.unece.org/display/ML/Machine+Learning+Group+2022}.

\bibitem[Chouldechova, 2016]{chouldechova_fair_2016}
Chouldechova, A. (2016).
\newblock Fair prediction with disparate impact: {A} study of bias in
  recidivism prediction instruments.
\newblock {\em arXiv:1610.07524 [cs, stat]}.

\bibitem[Christodoulou et~al.,
  2019]{Christodoulou.et.al.2019.logistic.regression.not.worse.than.machine.learning}
Christodoulou, E., Ma, J., Collins, G.~S., Steyerberg, E.~W., Verbakel, J.~Y.,
  and {Van Calster}, B. (2019).
\newblock {A systematic review shows no performance benefit of machine learning
  over logistic regression for clinical prediction models}.
\newblock {\em Journal of Clinical Epidemiology}, 110:12--22.

\bibitem[Clemmensen and Kj{\ae}rsgaard,
  2023]{clemmensen.2023.data.representativity.in.ml.and.ai}
Clemmensen, L.~H. and Kj{\ae}rsgaard, R.~D. (2023).
\newblock {Data Representativity for Machine Learning and AI Systems}.

\bibitem[Coronado and Ju{\'{a}}rez,
  2020]{coronado.juarez.2020.unece.imagery.theme.report}
Coronado, A. and Ju{\'{a}}rez, J. (2020).
\newblock {UNECE - HLG-MOS Machine Learning Project. Imagery Theme Report}.
\newblock v1, available at
  \url{https://statswiki.unece.org/display/ML/WP1+-+Theme+3+Imagery+Analysis+Report}.

\bibitem[Couper and Kreuter, 2013]{couper.kreuter.2013.response.times.jrssa}
Couper, M. and Kreuter, F. (2013).
\newblock Using paradata to explore item level response times in surveys.
\newblock {\em Journal of the Royal Statistical Society: Series A (Statistics
  in Society)}, 176(1):271--286.

\bibitem[Couper,
  2017]{couper.2017.new.developments.in.survey.data.collection.annual.review.in.soc}
Couper, M.~P. (2017).
\newblock {New Developments in Survey Data Collection}.
\newblock {\em Annual Review of Sociology}, 43:121--145.

\bibitem[Creel and Hellman, 2022]{creel.hellman.2022.algorithmic.leviathan}
Creel, K. and Hellman, D. (2022).
\newblock {The Algorithmic Leviathan: Arbitrariness, Fairness, and Opportunity
  in Algorithmic Decision-Making Systems}.
\newblock {\em Canadian Journal of Philosophy}, 52(1):26--43.

\bibitem[Curtin et~al., 2023]{cathal.et.al.2023.llms.in.official.stats}
Curtin, C., Senanayake, P., Clarke, C., Lichtenstein, I., Jamieson, A.,
  Roshanafshar, S., Yung, W., Piela, R., Vaiciulis, V., del Monaco, A.,
  Palumbo, L., Toepoel, V., Tingay, K., Banks, A., Bogdanova, B., Sirello, O.,
  Zdanowicz, K., Museux, J.-M., Tessitore, C., Danforth, J., Tebrake, J., Choi,
  I., and Kipkeeva, A. (2023).
\newblock {Large Language Models for Official Statistics}.
\newblock HLG-MOS White Paper, December 2023, available at
  \url{https://unece.org/sites/default/files/2023-12/HLGMOS%20LLM%20Paper_Preprint_1.pdf}.

\bibitem[Desiere et~al., 2019]{desiere_statistical_2019}
Desiere, S., Langenbucher, K., and Struyven, L. (2019).
\newblock Statistical profiling in public employment services: {An}
  international comparison.
\newblock {OECD} {Social}, {Employment} and {Migration} {Working} {Papers} 224,
  OECD Publishing, Paris, Franke.

\bibitem[Destatis, 2021]{destatis.2021.quality.manual}
Destatis (2021).
\newblock {\em {Quality Manual of the Statistical Offices of the Federation and
  the L{\"{a}}nder. (Original title: Qualit{\"{a}}tshandbuch der Statistischen
  {\"{A}}mter des Bundes und der L{\"{a}}nder)}}.
\newblock Wiesbaden, Germany.
\newblock
  \url{https://www.destatis.de/DE/Methoden/Qualitaet/qualitaetshandbuch.pdf}.

\bibitem[Doshi-Velez and Kim,
  2017]{doshivelez.kim.2017.rigorous.interpretable.ml.iml}
Doshi-Velez, F. and Kim, B. (2017).
\newblock {Towards A Rigorous Science of Interpretable Machine Learning}.

\bibitem[Doshi-Velez et~al.,
  2019]{doshivelez.et.al.2019.accountability.and.explanations.under.the.law}
Doshi-Velez, F., Kortz, M., Budish, R., Bavitz, C., Gershman, S., O'Brien, D.,
  Scott, K., Schieber, S., Waldo, J., Weinberger, D., Weller, A., and Wood, A.
  (2019).
\newblock Accountability of ai under the law: The role of explanation.

\bibitem[Dumpert, 2020]{dumpert.2020.unece.editing.imputation.theme.report}
Dumpert, F. (2020).
\newblock {UNECE - HLG-MOS Machine Learning Project. Edit and Imputation Theme
  Report}.
\newblock available at
  \url{https://statswiki.unece.org/display/ML/WP1+-+Theme+2+Edit+and+Imputation+Report}.

\bibitem[Dutta et~al.,
  2022]{dutta.et.al.2022.robust.counterfactual.explanations}
Dutta, S., Long, J., Mishra, S., Tilli, C., and Magazzeni, D. (2022).
\newblock {Robust Counterfactual Explanations for Tree-Based Ensembles}.
\newblock In Chaudhuri, K., Jegelka, S., Song, L., Szepesvari, C., Niu, G., and
  Sabato, S., editors, {\em Proceedings of the 39th International Conference on
  Machine Learning}, volume 162 of {\em Proceedings of Machine Learning
  Research}, pages 5742--5756. PMLR.

\bibitem[Dwork et~al., 2012]{dwork_fairness_2012}
Dwork, C., Hardt, M., Pitassi, T., Reingold, O., and Zemel, R. (2012).
\newblock Fairness through awareness.
\newblock In {\em Proceedings of the 3rd {Innovations} in {Theoretical}
  {Computer} {Science} {Conference} on - {ITCS} '12}, pages 214--226,
  Cambridge, Massachusetts. ACM Press.

\bibitem[Eckman,
  2013]{Eckman.2013.paradata.for.Coverage.Research.in.kreuter.paradata.book}
Eckman, S. (2013).
\newblock {Paradata for Coverage Research}.
\newblock In Kreuter, F., editor, {\em {Improving Surveys with Paradata:
  Analytic Uses of Process Information}}, pages 97--116. John Wiley \& Sons,
  Hoboken, NJ.

\bibitem[Engstrom et~al.,
  2020]{engstrom.et.al.2020.government.by.algorithm.ai.in.federal.agencies}
Engstrom, D.~F., Ho, D.~E., Sharkey, C.~M., and Cu{\'e}llar, M.-F., editors
  (2020).
\newblock {\em {Government by Algorithm: Artificial Intelligence in Federal
  Administrative Agencies}}, Public Law Research Paper 20-54. NYU School of
  Law.

\bibitem[{EU}, nd]{eu.nd.ai.watch}
{EU} (n.d.).
\newblock {AI Watch}.
\newblock Artificial intelligence website of the European Commission’s Joint
  Research Centre.

\bibitem[Eurostat, 2017]{eurostat.2017.eu.code.of.practice}
Eurostat (2017).
\newblock {European Statistics Code of Practice. Revised edition 2017}.

\bibitem[Fort, 2016]{fort.2016.collaborative.annotation.book}
Fort, K. (2016).
\newblock {\em {Collaborative Annotation for Reliable Natural Language
  Processing: Technical and Sociological Aspects}}.
\newblock {Wiley-ISTE}, Hoboken, NJ and London, UK.
\newblock \url{https://hal.science/hal-01324322}.

\bibitem[F{\"u}rnkranz et~al., 2012]{fuernkranz.et.a.2012.rule.learning}
F{\"u}rnkranz, J., Gamberger, D., and Lavra{\v{c}}, N. (2012).
\newblock {\em {Foundations of Rule Learning}}.
\newblock Springer, Heidelberg, Germany.

\bibitem[Gajane and Pechenizkiy, 2018]{gajane_formalizing_2018}
Gajane, P. and Pechenizkiy, M. (2018).
\newblock On {Formalizing} {Fairness} in {Prediction} with {Machine}
  {Learning}.
\newblock {\em arXiv:1710.03184 [cs, stat]}.

\bibitem[GCSILab, 2023]{athey.lab.2023.ml.causal.inference.tutorial}
GCSILab (2023).
\newblock {ML-based causal inference tutorial}.
\newblock last accessed 2023/04/08,
  \url{https://bookdown.org/stanfordgsbsilab/ml-ci-tutorial/}.

\bibitem[Gerdon et~al.,
  2022]{gerdon.et.al.2022.societal.impacts.of.adm.research.agenda.social.sciences}
Gerdon, F., Bach, R.~L., Kern, C., and Kreuter, F. (2022).
\newblock {Social impacts of algorithmic decision-making: A research agenda for
  the social sciences}.
\newblock {\em Big Data \& Society}, 9(1):1--13.

\bibitem[Ghani and Schierholz,
  2020]{ghani.schierholz.2020.machine.learning.ml.in.big.data.and.social.science.book}
Ghani, R. and Schierholz, M. (2020).
\newblock {Machine Learning}.
\newblock In Foster, I., Ghani, R., Jarmin, R.~S., Kreuter, F., and Lane, J.,
  editors, {\em {Big Data and Social Science}}, chapter~7. CRC Press, Boca
  Raton, FL, 2nd edition.

\bibitem[Goodman et~al., 2016]{goodman.et.al.2016.reproducibility.definitions}
Goodman, S.~N., Fanelli, D., and Ioannidis, J. P.~A. (2016).
\newblock What does research reproducibility mean?
\newblock {\em Science Translational Medicine}, 8(341):341ps12.

\bibitem[Grgic-Hlaca et~al., 2018]{grgic-hlaca_human_2018}
Grgic-Hlaca, N., Redmiles, E.~M., Gummadi, K.~P., and Weller, A. (2018).
\newblock Human {Perceptions} of {Fairness} in {Algorithmic} {Decision}
  {Making}: {A} {Case} {Study} of {Criminal} {Risk} {Prediction}.
\newblock In {\em Proceedings of the 2018 {World} {Wide} {Web} {Conference} on
  {World} {Wide} {Web} - {WWW} '18}, pages 903--912, Lyon, France. ACM Press.

\bibitem[Grinsztajn et~al.,
  2022]{Grinsztajn.et.al.2022.tree.based.outperform.deep.learning}
Grinsztajn, L., Oyallon, E., and Varoquaux, G. (2022).
\newblock {Why do tree-based models still outperform deep learning on tabular
  data?}

\bibitem[Groves, 2011]{groves.2011.3.eras.of.survey.research}
Groves, R.~M. (2011).
\newblock {Three Eras of Survey Research}.
\newblock {\em Public Opinion Quarterly}, 75(5):861--871.

\bibitem[Groves et~al., 2009]{groves.surveymeth}
Groves, R.~M., Fowler~Jr, F.~J., Couper, M.~P., Lepkowski, J.~M., Singer, E.,
  and Tourangeau, R. (2009).
\newblock {\em {Survey Methodology}}.
\newblock John Wiley \& Sons, Hoboken, NJ, 2nd edition.

\bibitem[Guts,
  2020]{guts.2020.workshop.on.target.leakage.in.ml.data.leakage.in.machine.learning}
Guts, Y. (2020).
\newblock {Workshop on Target Leakage in Machine Learning}.

\bibitem[Hampel et~al., 1986]{hampel.et.al.1986.robust.statistics}
Hampel, F.~R., Ronchetti, E.~M., Rousseeuw, P.~J., and Stahel, W.~A. (1986).
\newblock {\em {Robust Statistics: The Approach Based on Influence Functions}}.
\newblock John Wiley \& Sons, Hoboken, NJ.

\bibitem[Hastie et~al., 2009]{Hastie2009}
Hastie, T., Tibshirani, R., and Friedman, J. (2009).
\newblock {\em The Elements of Statistical Learning: Data Mining, Inference,
  and Prediction}.
\newblock New York, NY: Springer.

\bibitem[Hebert-Johnson et~al., 2018]{hebert-johnson_multicalibration_2018}
Hebert-Johnson, U., Kim, M., Reingold, O., and Rothblum, G. (2018).
\newblock Multicalibration: {Calibration} for the
  ({Computationally}-{Identifiable}) {Masses}.
\newblock In Dy, J. and Krause, A., editors, {\em Proceedings of the 35th
  {International} {Conference} on {Machine} {Learning}}, volume~80 of {\em
  Proceedings of {Machine} {Learning} {Research}}, pages 1939--1948,
  Stockholmsm{\"{a}}ssan, Stockholm, Sweden. PMLR.

\bibitem[Heidari et~al., 2019]{heidari_moral_2019}
Heidari, H., Loi, M., Gummadi, K.~P., and Krause, A. (2019).
\newblock A {Moral} {Framework} for {Understanding} {Fair} {ML} through
  {Economic} {Models} of {Equality} of {Opportunity}.
\newblock In {\em Proceedings of the {Conference} on {Fairness},
  {Accountability}, and {Transparency}}, pages 181--190, Atlanta GA USA. ACM.

\bibitem[Helwegen and Braaksma,
  2020]{helwegen.braaksma.2020.fair.algorithms.in.context.netherlands.center.for.big.data.stats}
Helwegen, R. and Braaksma, B. (2020).
\newblock Fair algorithms in context.
\newblock Working paper no. 05-20.

\bibitem[Herzog et~al.,
  2007]{herzog.scheuren.winkler.2007.data.quality.record.linkage}
Herzog, T.~N., Scheuren, F.~J., and Winkler, W.~E. (2007).
\newblock {\em {Data Quality and Record Linkage Techniques}}.
\newblock Springer, New York, NY.

\bibitem[Hill et~al., 2019]{hill.et.al.2019.bigsurv18}
Hill, C.~A., Biemer, P., Buskirk, T., Callegaro, M., C{\'o}rdova~Cazar, A.~L.,
  Eck, A., Japec, L., Kirchner, A., Kolenikov, S., Lyberg, L., and Sturgis, P.
  (2019).
\newblock {Exploring New Statistical Frontiers at the Intersection of Survey
  Science and Big Data: Convergence at ``BigSurv18''}.
\newblock {\em Survey Research Methods}, 13(1):123--135.

\bibitem[Hill et~al.,
  2021]{hill.et.al.2021.big.data.meets.survey.science.methods}
Hill, C.~A., Biemer, P.~P., Buskirk, T.~D., Japec, L., Kirchner, A., Kolenikov,
  S., and Lyberg, L.~E. (2021).
\newblock {\em {Big Data Meets Survey Science: A Collection of Innovative
  Methods}}.
\newblock John Wiley \& Sons, Hoboken, NJ.

\bibitem[Holloway and Mengersen,
  2018]{holloway.mengersen.2018.remote.sensing.for.SDGs}
Holloway, J. and Mengersen, K. (2018).
\newblock Statistical machine learning methods and remote sensing for
  sustainable development goals: A review.
\newblock {\em Remote Sensing}, 10(9).

\bibitem[Hornik, 2005]{hornik.2005.clue.explaination}
Hornik, K. (2005).
\newblock A {CLUE} for {CLUster Ensembles}.
\newblock {\em Journal of Statistical Software}, 14(12).

\bibitem[Hou and Jung,
  2021]{hou.jung.2021.reconciling.algorithm.aversion.and.appreciation}
Hou, Y. T.-Y. and Jung, M.~F. (2021).
\newblock {Who is the Expert? Reconciling Algorithm Aversion and Algorithm
  Appreciation in AI-Supported Decision Making}.
\newblock {\em Proceedings of the ACM on Human-Computer Interaction},
  5(CSCW2):1--25.

\bibitem[Huber and Ronchetti, 2009]{huber.ronchetti.2009.robust.statistics}
Huber, P.~J. and Ronchetti, E.~M. (2009).
\newblock {\em {Robust Statistics}}.
\newblock John Wiley \& Sons, Hoboken, NJ.

\bibitem[{IPS Observatory}, nd]{ips.nd.database.of.ai.in.public.services}
{IPS Observatory} (n.d.).
\newblock {IPS-X. The Innovative Public Services Explorer}.

\bibitem[James et~al., 2021]{james.et.al.2021.intro.statistical.learning}
James, G., Witten, D., Hastie, T., and Tibshirani, R. (2021).
\newblock {\em {An Introduction to Statistical Learning}}.
\newblock Springer, New York, NY, 2nd edition.
\newblock First printing August 4, 2021, pdf accessed August 31, 2021.

\bibitem[Japec et~al.,
  2015]{japec.kreuter.biemer.lane.et.al.2015.big.data.survey.research.aapor.task.force.report}
Japec, L., Kreuter, F., Berg, M., Biemer, P.~P., Decker, P., Lampe, C., Lane,
  J., O'Neil, C., and Usher, A. (2015).
\newblock {Big Data in Survey Research: AAPOR Task Force Report}.
\newblock {\em Public Opinion Quarterly}, 79(4):839--880.

\bibitem[Julien, 2020]{julien.2020.unece.hlg.mos.ml.project.report}
Julien, C. (2020).
\newblock {UNECE -- HLG-MOS Machine Learning Project Project report}.
\newblock v2, available at
  \url{https://statswiki.unece.org/display/ML/Machine+Learning+Project+Report}.

\bibitem[Jussupow et~al.,
  2020]{jussupow.et.al.2020.algorithm.aversion.literature.review}
Jussupow, E., Benbasat, I., and Heinzl, A. (2020).
\newblock {Why are we averse towards algorithms? A comprehensive literature
  review on algorithm aversion}.
\newblock In {\em Proceedings of the 28th European Conference on Information
  Systems (ECIS), An Online AIS Conference}.

\bibitem[Kaiser et~al., 2022]{kaiser2022}
Kaiser, P., Kern, C., and R{\"{u}}gamer, D. (2022).
\newblock Uncertainty-aware predictive modeling for fair data-driven decisions.

\bibitem[Kapoor and Narayanan,
  2022]{kapoor.narayanan.2022.leakage.reproducibility.crisis.in.ML}
Kapoor, S. and Narayanan, A. (2022).
\newblock {Leakage and the Reproducibility Crisis in ML-based Science}.

\bibitem[Karimi et~al., 2021]{karimi.et.al.2021.algorithmic.recourse.survey}
Karimi, A.-H., Barthe, G., Sch{\"{o}}lkopf, B., and Valera, I. (2021).
\newblock A survey of algorithmic recourse: definitions, formulations,
  solutions, and prospects.

\bibitem[Kearns et~al., 2018]{kearns2018}
Kearns, M., Neel, S., Roth, A., and Wu, Z.~S. (2018).
\newblock Preventing fairness gerrymandering: Auditing and learning for
  subgroup fairness.
\newblock In Dy, J. and Krause, A., editors, {\em Proceedings of the 35th
  International Conference on Machine Learning}, volume~80 of {\em Proceedings
  of Machine Learning Research}, pages 2564--2572. PMLR.

\bibitem[Kern et~al., 2021]{kern2021}
Kern, C., Bach, R., Mautner, H., and Kreuter, F. (2021).
\newblock Fairness in algorithmic profiling: A german case study.
\newblock {\em arXiv:2108.04134}.
\newblock \url{https://arxiv.org/pdf/2108.04134}.

\bibitem[Kern et~al., 2022]{Kern2022}
Kern, C., Gerdon, F., Bach, R.~L., Keusch, F., and Kreuter, F. (2022).
\newblock {Humans versus machines: Who is perceived to decide fairer?
  Experimental evidence on attitudes toward automated decision-making}.
\newblock {\em Patterns}, 3(10):100591.

\bibitem[Keusch et~al., tion]{keusch.et.al.wearables.sensors.apps}
Keusch, F., Struminskaya, B., Eckman, S., and Guyer, H.~M. ({in preparation}).
\newblock {\em {Data Collection with Wearables, Apps, and Sensors}}.
\newblock CRC Press.

\bibitem[Kilbertus et~al., 2017]{Kilbertus2017}
Kilbertus, N., Rojas-Carulla, M., Parascandolo, G., Hardt, M., Janzing, D., and
  Sch\"{o}lkopf, B. (2017).
\newblock Avoiding discrimination through causal reasoning.
\newblock In {\em Proceedings of the 31st International Conference on Neural
  Information Processing Systems}, NIPS'17, page 656–666, Red Hook, NY, USA.
  Curran Associates Inc.

\bibitem[Kim et~al., 2019]{kim_multiaccuracy_2019}
Kim, M.~P., Ghorbani, A., and Zou, J. (2019).
\newblock Multiaccuracy: {Black}-{Box} {Post}-{Processing} for {Fairness} in
  {Classification}.
\newblock In {\em Proceedings of the 2019 {AAAI}/{ACM} {Conference} on {AI},
  {Ethics}, and {Society}}, {AIES} '19, pages 247--254, New York, NY, USA.
  Association for Computing Machinery.
\newblock event-place: Honolulu, HI, USA.

\bibitem[Kleinberg and Raghavan,
  2021]{kleinberg.raghavan.2021.algorithmic.monoculture}
Kleinberg, J. and Raghavan, M. (2021).
\newblock {Algorithmic monoculture and social welfare}.
\newblock {\em Proceedings of the National Academy of Sciences},
  118(22):e2018340118.

\bibitem[K{\"o}rtner and Bonoli, 2021]{kortner2021}
K{\"o}rtner, J. and Bonoli, G. (2021).
\newblock {Predictive Algorithms in the Delivery of Public Employment
  Services}.
\newblock {\em SocArXiv}.
\newblock \url{https://osf.io/j7r8y/}.

\bibitem[Kreuter, 2013]{fkpara}
Kreuter, F., editor (2013).
\newblock {\em {Improving Surveys with Paradata: Analytic Uses of Process
  Information}}, Hoboken, NJ. John Wiley \& Sons.

\bibitem[Krishna et~al.,
  2022]{krishna.lakkaraju.et.al.2022.disagreement.in.XAI.practitioner.perspective}
Krishna, S., Han, T., Gu, A., Pombra, J., Jabbari, S., Wu, S., and Lakkaraju,
  H. (2022).
\newblock {The Disagreement Problem in Explainable Machine Learning: A
  Practitioner's Perspective}.

\bibitem[Kuppler et~al., 2022]{kuppler2022fair}
Kuppler, M., Kern, C., Bach, R., and Kreuter, F. (2022).
\newblock From fair predictions to just decisions? conceptualizing algorithmic
  fairness and distributive justice in the context of data-driven
  decision-making.
\newblock {\em Frontiers in Sociology}, 7.

\bibitem[Kusner et~al., 2018]{kusner_counterfactual_2018}
Kusner, M.~J., Loftus, J.~R., Russell, C., and Silva, R. (2018).
\newblock Counterfactual {Fairness}.
\newblock {\em arXiv:1703.06856 [cs, stat]}.

\bibitem[Lakkaraju et~al.,
  2022]{lakkaraju.et.al.2022.explainability.practitioner.dialogue}
Lakkaraju, H., Slack, D., Chen, Y., Tan, C., and Singh, S. (2022).
\newblock {Rethinking Explainability as a Dialogue: A Practitioner's
  Perspective}.

\bibitem[Lee et~al., 2020]{lee_fairness_2020}
Lee, M. S.~A., Floridi, L., and Singh, J. (2020).
\newblock From {Fairness} {Metrics} to {Key} {Ethics} {Indicators} ({KEIs}):
  {A} {Context}-{Aware} {Approach} to {Algorithmic} {Ethics} in an {Unequal}
  {Society}.
\newblock {\em SSRN Electronic Journal}.

\bibitem[Ligozat et~al.,
  2022]{ligozat.et.al.2022.unraveling.environmental.impact.ai}
Ligozat, A.-L., Lef{\`{e}}vre, J., Bugeau, A., and Combaz, J. (2022).
\newblock {Unraveling the Hidden Environmental Impacts of AI Solutions for
  Environment}.

\bibitem[Little and Rubin,
  2019]{little.2019.statistical.analysis.with.missing.data}
Little, R.~J. and Rubin, D.~B. (2019).
\newblock {\em Statistical analysis with missing data}.
\newblock John Wiley \& Sons, Hoboken, NJ.

\bibitem[Loi et~al., 2021]{loi_fair_2021}
Loi, M., Herlitz, A., and Heidari, H. (2021).
\newblock Fair {Equality} of {Chances} for {Prediction}-{Based} {Decisions}.
\newblock In {\em Proceedings of the 2021 {AAAI}/{ACM} {Conference} on {AI},
  {Ethics}, and {Society}}, {AIES} '21, page 756, New York, NY, USA.
  Association for Computing Machinery.
\newblock event-place: Virtual Event, USA.

\bibitem[Makhlouf et~al., 2020]{makhlouf_applicability_2020}
Makhlouf, K., Zhioua, S., and Palamidessi, C. (2020).
\newblock On the {Applicability} of {ML} {Fairness} {Notions}.
\newblock {\em arXiv:2006.16745 [cs, stat]}.

\bibitem[Makhlouf et~al., 2022]{makhlouf.et.al.2022.survey.causal.fair.ml}
Makhlouf, K., Zhioua, S., and Palamidessi, C. (2022).
\newblock {Survey on Causal-based Machine Learning Fairness Notions}.

\bibitem[Measure, 2020]{measure.2020.unece.hlg.mos.ml.integration}
Measure, A. (2020).
\newblock {UNECE -- HLG-MOS Machine Learning Project. Work Package 3 -
  Integration}.
\newblock v0.4 final, available at
  \url{https://statswiki.unece.org/display/ML/WP3+-+Integration}.

\bibitem[Mehrabi et~al., 2021]{mehrabi_survey_2021}
Mehrabi, N., Morstatter, F., Saxena, N., Lerman, K., and Galstyan, A. (2021).
\newblock A survey on bias and fairness in machine learning.
\newblock {\em ACM Computing Surveys}, 54(6).

\bibitem[Miller,
  2017]{miller.2017.Explanation.in.AI.Insights.from.Social.Sciences}
Miller, T. (2017).
\newblock {Explanation in Artificial Intelligence: Insights from the Social
  Sciences}.

\bibitem[Mitchell et~al.,
  2019]{mitchell.et.al.gebru.2019.model.cards.for.model.reporting}
Mitchell, M., Wu, S., Zaldivar, A., Barnes, P., Vasserman, L., Hutchinson, B.,
  Spitzer, E., Raji, I.~D., and Gebru, T. (2019).
\newblock {Model Cards for Model Reporting}.
\newblock In {\em Proceedings of the Conference on Fairness, Accountability,
  and Transparency}. {ACM}.

\bibitem[Mitchell et~al., 2021]{mitchell_algorithmic_2021}
Mitchell, S., Potash, E., Barocas, S., D'Amour, A., and Lum, K. (2021).
\newblock Algorithmic {Fairness}: {Choices}, {Assumptions}, and {Definitions}.
\newblock {\em Annual Review of Statistics and Its Application}, 8(1):141--163.

\bibitem[Mittereder,
  2019]{mittereder.2019.predicting.preventing.breakoff.web.surveys.phd.thesis}
Mittereder, F.~K. (2019).
\newblock {\em {Predicting and Preventing Breakoff in Web Surveys}}.
\newblock Dissertation, University of Michigan, Ann Arbor, MI.

\bibitem[Molnar, 2020]{molnar.2020.interpretable.ml}
Molnar, C. (2020).
\newblock {\em {Interpretable Machine Learning. A Guide for Making Black Box
  Models Explainable}}.
\newblock \href{https://leanpub.com/u/ChristophMolnar}{Leanpub}, 2nd edition.
\newblock
  \href{https://christophm.github.io/interpretable-ml-book}{christophm.github.io/interpretable-ml-book},
  2nd edition 2022.

\bibitem[Molnar, 2022]{molnar.2022.modeling.mindsets}
Molnar, C. (2022).
\newblock {\em {Modeling Mindsets. The Many Cultures of Learning From Data}}.
\newblock Independently published at
  \href{https://leanpub.com/u/ChristophMolnar}{Leanpub}.
\newblock \url{www.modeling-mindsets.com}.

\bibitem[Moreno-Torres et~al., 2012]{morreno.torres.et.al.2012.dataset.shift}
Moreno-Torres, J.~G., Raeder, T., Alaiz-Rodríguez, R., Chawla, N.~V., and
  Herrera, F. (2012).
\newblock A unifying view on dataset shift in classification.
\newblock {\em Pattern Recognition}, 45(1):521--530.

\bibitem[Murdoch et~al., 2019]{murdoch.et.al.2019.interpretable.ml}
Murdoch, W.~J., Singh, C., Kumbier, K., Abbasi-Asl, R., and Yu, B. (2019).
\newblock {Definitions, methods, and applications in interpretable machine
  learning}.
\newblock {\em Proceedings of the National Academy of Sciences},
  116(44):22071--22080.

\bibitem[Neunhoeffer et~al.,
  2021]{neunhoeffer.et.al.2021.private.generative.adversarial.networks}
Neunhoeffer, M., Wu, Z.~S., and Dwork, C. (2021).
\newblock {Private Post-GAN Boosting}.

\bibitem[Obermeyer et~al., 2019]{obermeyer_dissecting_2019}
Obermeyer, Z., Powers, B., Vogeli, C., and Mullainathan, S. (2019).
\newblock Dissecting racial bias in an algorithm used to manage the health of
  populations.
\newblock {\em Science}, 366(6464):447--453.

\bibitem[Ohme et~al., 2023]{ohme.et.al.2023.api.data.donation.tracking}
Ohme, J., Araujo, T., Boeschoten, L., Freelon, D., Ram, N., Reeves, B.~B., and
  Robinson, T.~N. (2023).
\newblock {Digital Trace Data Collection for Social Media Effects Research:
  APIs, Data Donation, and (Screen) Tracking}.
\newblock {\em Communication Methods and Measures}, pages 1--18.
\newblock Published online 2023/02/27.

\bibitem[Perdomo et~al., 2020]{Perdomo2020}
Perdomo, J.~C., Zrnic, T., Mendler-D{\"{u}}nner, C., and Hardt, M. (2020).
\newblock {Performative Prediction}.

\bibitem[Plecko and Bareinboim,
  2022]{plecko.barenboim.2022.causal.fairness.analysis}
Plecko, D. and Bareinboim, E. (2022).
\newblock {Causal Fairness Analysis}.

\bibitem[Plesser,
  2018]{plesser.2018.reproducibility.replicability.history.of.terminology}
Plesser, H.~E. (2018).
\newblock {Reproducibility vs. Replicability: A Brief History of a Confused
  Terminology}.
\newblock {\em Frontiers in Neuroinformatics}, 11.

\bibitem[Puts et~al.,
  2022]{unece.2022.quality.of.training.data.theme.group.report}
Puts, M. J.~H., da~Silva, A., Di~Consiglio, L., Choi, I., Salgado, D., Clarke,
  C., Jones, S., and Baily, A. (2022).
\newblock {ONS-UNECE Machine Learning Group 2022. Quality of training data.
  Theme Group Report}.
\newblock v1, available at
  \url{https://statswiki.unece.org/display/ML/Machine+Learning+Group+2022}.

\bibitem[{Qui{\~{n}}onero-Candela} et~al.,
  2008]{quinonero.candela.et.al.2008.dataset.shift.in.ml.edited.volume}
{Qui{\~{n}}onero-Candela}, J., Sugiyama, M., Schwaighofer, A., and Lawrence,
  N.~D., editors (2008).
\newblock {\em {Dataset Shift in Machine Learning}}, Cambridge, MA. MIT Press.

\bibitem[Reusens et~al., 2022]{unece.2022.web.scraping.theme.group.report}
Reusens, M., Kurban, B., Peszat, K., Grancow, B., and Murawska, E. (2022).
\newblock {ML2022: Web scraping theme group report}.
\newblock v1, available at
  \url{https://statswiki.unece.org/display/ML/Machine+Learning+Group+2022}.

\bibitem[Richards et~al., 2020]{richards.et.al.2020.creating.ai.factsheets}
Richards, J., Piorkowski, D., Hind, M., Houde, S., and Mojsilovi{\'c}, A.
  (2020).
\newblock {A Methodology for Creating AI FactSheets}.

\bibitem[Roberts et~al.,
  2021]{roberts.et.al.2021.common.pitfalls.covid.ml.models}
Roberts, M., Driggs, D., Thorpe, M., Gilbey, J., Yeung, M., Ursprung, S.,
  Aviles-Rivero, A.~I., Etmann, C., McCague, C., Beer, L., et~al. (2021).
\newblock {Common pitfalls and recommendations for using machine learning to
  detect and prognosticate for COVID-19 using chest radiographs and CT scans}.
\newblock {\em Nature Machine Intelligence}, 3(3):199--217.

\bibitem[Rodolfa et~al., 2020]{rodolfa_bias_2021}
Rodolfa, K.~T., Saleiro, P., and Ghani, R. (2020).
\newblock Bias and {Fairness}.
\newblock In Foster, I., Ghani, R., Jarmin, R.~S., Kreuter, F., and Lane, J.,
  editors, {\em {Big Data and Social Science}}, chapter~7. CRC Press, Boca
  Raton, FL, 2nd edition.

\bibitem[Saleiro et~al., 2019]{saleiro_aequitas_2019}
Saleiro, P., Kuester, B., Hinkson, L., London, J., Stevens, A., Anisfeld, A.,
  Rodolfa, K.~T., and Ghani, R. (2019).
\newblock {Aequitas: A Bias and Fairness Audit Toolkit}.

\bibitem[Salwiczek and Rohde,
  2022]{salwiczek.rohde.2022.quality.in.official.statistics.workshop.presentation}
Salwiczek, C. and Rohde, J. (2022).
\newblock {Dimensions of Quality for the Use of ML in Official Statistics}.
\newblock Presented at the Workshop ``Quality Aspects of Machine Learning --
  Official Statistics between Specific Quality Requirements and Methodological
  Innovation, Munich, Germany, 2022/09/06.

\bibitem[Schenk and Reu{\ss}, 2023]{schenk.reuss.2023.springer.volume.paradata}
Schenk, P. and Reu{\ss}, S. (2023).
\newblock {Paradata in Surveys}.
\newblock In Huvila, I., B{\"o}rjesson, L., Kaiser, J., Friberg, Z., and
  Sk{\"o}ld, O., editors, {\em {Perspectives to Paradata}}. Springer.

\bibitem[Schwanh{\"a}user et~al.,
  2022]{schwanhaeuser.2022.interviewer.falsification}
Schwanh{\"a}user, S., Sakshaug, J.~W., and Kosyakova, Y. (2022).
\newblock {How to Catch a Falsifier: Comparison of Statistical Detection
  Methods for Interviewer Falsification}.
\newblock {\em Public Opinion Quarterly}, 86(1):51--81.

\bibitem[Schwartz et~al., 2019]{schwartz.et.al.2019.green.ai}
Schwartz, R., Dodge, J., Smith, N.~A., and Etzioni, O. (2019).
\newblock {Green AI}.

\bibitem[Seibold,
  2023]{seibold.2023.quote.open.science.good.science.digital.age}
Seibold, H. (2023).
\newblock {Newsletter, 2023/04/21}.

\bibitem[Srivastava et~al., 2019]{Srivastava2019}
Srivastava, M., Heidari, H., and Krause, A. (2019).
\newblock Mathematical notions vs. human perception of fairness: A descriptive
  approach to fairness for machine learning.
\newblock In {\em Proceedings of the 25th ACM SIGKDD International Conference
  on Knowledge Discovery \& Data Mining}, KDD '19, pages 2459--2468, New York,
  NY, USA. Association for Computing Machinery.

\bibitem[Starke et~al., 2022]{Starke2022}
Starke, C., Baleis, J., Keller, B., and Marcinkowski, F. (2022).
\newblock Fairness perceptions of algorithmic decision-making: A systematic
  review of the empirical literature.
\newblock {\em Big Data \& Society}, 9(2).

\bibitem[Sthamer, 2020a]{sthamer.2020.editing.social.survey.data.with.ML}
Sthamer, C. (2020a).
\newblock {Editing of Social Survey Data with Machine Learning -- A journey
  from PoC to Implementation}.
\newblock v2, 2022-10-15, available at
  \url{https://statswiki.unece.org/display/ML/Editing+of+Social+Survey+Data+with+Machine+Learning+-+A+journey+from+PoC+to+Implementation}.

\bibitem[Sthamer, 2020b]{sthamer.2020.classification.coding.theme.report}
Sthamer, C. (2020b).
\newblock {UNECE -- HLG-MOS Machine Learning Project. Classification and Coding
  Theme Report}.
\newblock v6, available at
  \url{https://statswiki.unece.org/display/ML/WP1+-+Theme+1+Coding+and+Classification+Report}.

\bibitem[{Text Classification Theme Group},
  2022]{unece.2022.text.classification.theme.group.report}
{Text Classification Theme Group} (2022).
\newblock {ML 2022 Text Classification Theme Group Report}.
\newblock v1, available at
  \url{https://statswiki.unece.org/display/ML/Machine+Learning+Group+2022}.

\bibitem[Tokle and Bender,
  2020]{tokle.bender.2020.record.linkage.in.big.data.and.social.science.book}
Tokle, J. and Bender, S. (2020).
\newblock {Record Linkage}.
\newblock In Foster, I., Ghani, R., Jarmin, R.~S., Kreuter, F., and Lane, J.,
  editors, {\em {Big Data and Social Science}}, chapter~§. CRC Press, Boca
  Raton, FL, 2nd edition.

\bibitem[Tornede et~al., 2022]{tornede.et.al.2022green.auto.ml}
Tornede, T., Tornede, A., Hanselle, J., Wever, M., Mohr, F., and
  H{\"{u}}llermeier, E. (2022).
\newblock {Towards Green Automated Machine Learning: Status Quo and Future
  Directions}.

\bibitem[Tourangeau et~al.,
  2014]{tourangeau.et.al.2014.hard.to.survey.populations.edited.volume}
Tourangeau, R., Edwards, B., Johnson, T.~P., Wolter, K.~M., and Bates, N.,
  editors (2014).
\newblock {\em {Hard-to-Survey Populations}}, Cambridge, UK. Cambridge
  University Press.

\bibitem[TrustML, nd]{trust.ml.nd.homepage}
TrustML (n.d.).
\newblock {The Trustworthy ML Initiative}.
\newblock https://www.trustworthyml.org/resources.

\bibitem[{UK Statistics Authority},
  2021]{unece.2021.ethical.considerations.of.ml}
{UK Statistics Authority} (2021).
\newblock {Ethical considerations in the use of Machine Learning for research
  and statistics}.
\newblock ONS-UNECE Machine Learning Group 2021 Work Stream 3, available at
  \url{https://uksa.statisticsauthority.gov.uk/publication/ethical-considerations-in-the-use-of-machine-learning-for-research-and-statistics/pages/1/}.

\bibitem[UNECE, 2013]{unece.2013.fundamental.principles.official.statistics}
UNECE (2013).
\newblock {Fundamental Principles of Official Statistics}.
\newblock Revised version, adopted on 24 July 2013.

\bibitem[Varshney, 2022]{varshney.2022.trustworthy.ml.book}
Varshney, K.~R. (2022).
\newblock {\em Trustworthy Machine Learning}.
\newblock Independently Published, Chappaqua, NY.
\newblock
  \href{http://www.trustworthymachinelearning.com}{trustworthymachinelearning.com}.

\bibitem[Verma et~al.,
  2020]{verma.et.al.2020.counterfactual.explanations.algorithmic.recourse.literate.review}
Verma, S., Boonsanong, V., Hoang, M., Hines, K.~E., Dickerson, J.~P., and Shah,
  C. (2020).
\newblock {Counterfactual Explanations and Algorithmic Recourses for Machine
  Learning: A Review}.
\newblock v3, 15 Nov 2022.

\bibitem[von Zahn et~al., 2023]{vonzahn2023locating}
von Zahn, M., Hinz, O., and Feuerriegel, S. (2023).
\newblock Locating disparities in machine learning.

\bibitem[Wachter et~al.,
  2017a]{wachter.et.al.2017.no.right.to.explanation.in.gdpr}
Wachter, S., Mittelstadt, B., and Floridi, L. (2017a).
\newblock {Why a Right to Explanation of Automated Decision-Making Does Not
  Exist in the General Data Protection Regulation}.
\newblock {\em International Data Privacy Law}, 7(2):76--99.

\bibitem[Wachter et~al.,
  2017b]{wachter.et.al.2017.counterfactual.explanations.without.opening.the.black.box}
Wachter, S., Mittelstadt, B., and Russell, C. (2017b).
\newblock Counterfactual explanations without opening the black box: Automated
  decisions and the gdpr.
\newblock {\em Harvard Journal of Law \& Technology}, 31(2):841--887.

\bibitem[Wagner, 2008]{wagner.2008.Adaptive.Survey.Design.phd.thesis}
Wagner, J.~R. (2008).
\newblock {\em {Adaptive Survey Design to Reduce Nonresponse Bias}}.
\newblock Dissertation, University of Michigan, Ann Arbor, MI.

\bibitem[Weerts et~al., 2023a]{weerts2023}
Weerts, H., Pfisterer, F., Feurer, M., Eggensperger, K., Bergman, E., Awad, N.,
  Vanschoren, J., Pechenizkiy, M., Bischl, B., and Hutter, F. (2023a).
\newblock {Can Fairness be Automated? Guidelines and Opportunities for
  Fairness-aware AutoML}.

\bibitem[Weerts et~al., 2023b]{weerts.et.al.2023.fairness.auto.ml}
Weerts, H., Pfisterer, F., Feurer, M., Eggensperger, K., Bergman, E., Awad, N.,
  Vanschoren, J., Pechenizkiy, M., Bischl, B., and Hutter, F. (2023b).
\newblock {Can Fairness be Automated? Guidelines and Opportunities for
  Fairness-aware AutoML}.

\bibitem[West et~al., 2023]{west.wagner.2023.tdq}
West, B.~T., Wagner, J., Kim, J., and Buskirk, T.~D. (2023).
\newblock {The Total Data Quality Framework}.
\newblock \url{https://www.coursera.org/specializations/total-data-quality}.
\newblock Online; accessed 13 March 2023.

\bibitem[Willis et~al.,
  2014]{willis.et.al.2014.overview.special.issue.hard.to.reach}
Willis, G.~B., Smith, T.~W., Shariff-Marco, S., and English, N. (2014).
\newblock {Overview of the Special Issue on Surveying the Hard-to-Reach}.
\newblock {\em Journal of Official Statistics}, 30(2):171--176.

\bibitem[Yung et~al.,
  2022]{yung.et.al.2022.quality.framework.statistical.algorithms}
Yung, W., Tam, S.-M., Buelens, B., Chipman, H., Dumpert, F., Ascari, G., Rocci,
  F., Burger, J., and Choi, I. (2022).
\newblock {A Quality Framework for Statistical Algorithms}.
\newblock {\em Statistical Journal of the IAOS}, 38(1):291--308.
\newblock Referenced page numbers refer to preprint available at
  \url{https://statswiki.unece.org/download/attachments/285216420/QF4SA_2020_Final.pdf}.

\end{thebibliography}
